\documentclass{article} 
\usepackage{arxiv,times}

\usepackage{url}
\usepackage{hyperref}
\usepackage{multirow}
\usepackage{multicol}
\usepackage{subcaption}
\usepackage{wrapfig}
\usepackage{caption}
\usepackage{graphicx}
\usepackage{algorithm}
\usepackage{algpseudocode}
\usepackage{amssymb}
\usepackage{booktabs}
\usepackage{adjustbox}
\usepackage{mathtools}
\usepackage{color}
\usepackage{array}
\usepackage{caption}
\usepackage{pifont}

\title{STBLLM: Breaking the 1-Bit Barrier with \\Structured Binary LLMs}

\author{
Peijie Dong$^{1,\dag}$, 
Lujun Li$^{2,\dag}$, 
Yuedong Zhong$^{3}$, 
Dayou Du$^{1}$, 
Ruibo Fan$^{1}$, 
Yuhan Chen$^{1}$, \\
\textbf{Zhenheng Tang}$^{1,4}$, 
\textbf{Qiang Wang}$^{5}$, 
\textbf{Wei Xue}$^{2}$, 
\textbf{Yike Guo}$^{2,*}$, 
\textbf{Xiaowen Chu}$^{1,2}$\thanks{Corresponding authors. $\dag$ Equal contribution.} \\
$^1$ HKUST(GZ) \quad 
$^2$ HKUST \quad 
$^3$ SYSU \quad 
$^4$ HKBU \quad 
$^5$ HIT(SZ) 
\\
{\tt\small pdong212@connect.hkust-gz.edu.cn, lilujunai@gmail.com,}\\
{\tt\small zhongyd6@mail2.sysu.edu.cn, \{ddu487, rfan404, ychen906\}@connect.hkust-gz.edu.cn,}\\
{\tt\small zhtang@comp.hkbu.edu.hk, qiang.wang@hit.edu.cn, \{yikeguo, xwchu\}@ust.hk}\\
}

\vspace{-2.5em}

\iclrfinalcopy
\begin{document}

\maketitle

\begin{abstract}
In this paper, we present the first structural binarization method for LLM compression to less than 1-bit precision. Although LLMs have achieved remarkable performance, their memory-bound nature during the inference stage hinders the adoption of resource-constrained devices. Reducing weights to 1-bit precision through binarization substantially enhances computational efficiency. We observe that some weights in binarized LLMs can be randomly flipped without significant performance degradation, suggesting the potential for further compression. To exploit this, our STBLLM employs an N:M sparsity technique to achieve structural binarization of the weights. Specifically, we introduce a novel Standardized Importance (SI) metric, which considers weight magnitude and input feature norm to more accurately assess weight significance. Then, we propose a layer-wise approach, allowing different layers of the LLM to be sparsified with varying N:M ratios, thereby balancing compression and accuracy. Furthermore, we implement a fine-grained grouping strategy for less important weights, applying distinct quantization schemes to sparse, intermediate, and dense regions. Finally, we design a specialized CUDA kernel to support structural binarization. We conduct extensive experiments on LLaMA-1/2/3, OPT family, and Mistral to evaluate the effectiveness of STBLLM. The results demonstrate that our approach performs better than other compressed binarization LLM methods while significantly reducing memory requirements.

\end{abstract}

\section{Introduction}

The advent of large language models (LLMs), such as \citep{Zhang2022OPTOP, touvron2023llama, brown2020language}, has revolutionized the field of natural language processing (NLP) \citep{wei2022chain}. These powerful models exhibit remarkable performance, surpassing human capabilities in certain domains \citep{wei2022emergent, bubeck2023sparks}. However, the immense scale and complexity of LLMs present significant challenges in terms of memory requirements, hindering their widespread deployment, especially in resource-constrained environments. 
To address this issue, model compression techniques, such as quantization~\citep{Frantar2022GPTQAP, Lin2024DuQuantDO, dong2023emq}, pruning~\citep{meng2020pruning}, distillation~\citep{gu2023knowledge}, and low-rank decomposition~\citep{ashkboos2024slicegpt}, have gained increasing attention in reducing the computational footprint while preserving their performance. One promising approach is network binarization, the most aggressive quantization method. Binarization quantizes original floating-point weights with binary values  ($-1$ or $+1$), significantly reduces memory storage.

\begin{figure}[t]
    \centering
    \begin{minipage}{0.48\textwidth}
        \centering
        \includegraphics[width=\linewidth]{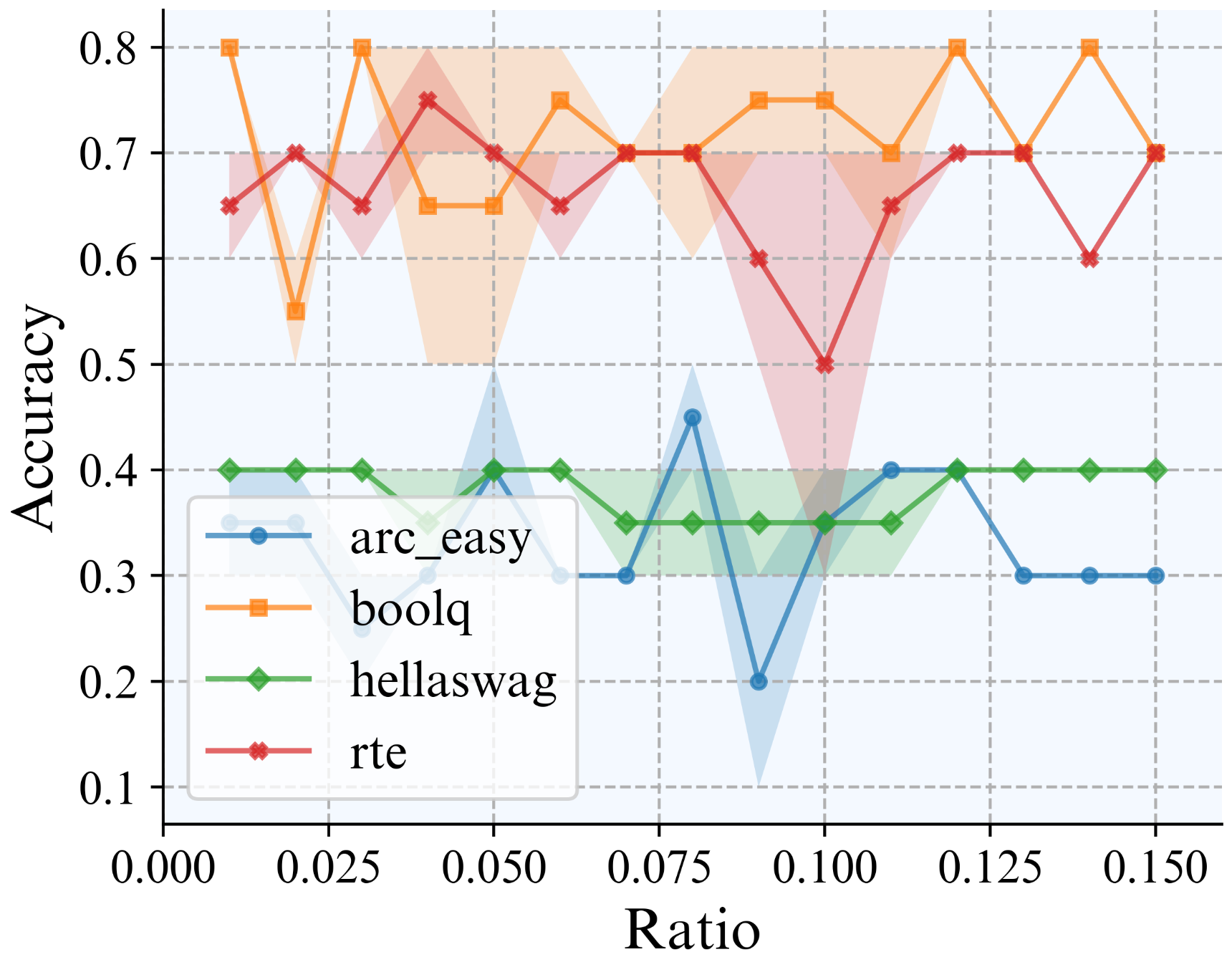}
        \vspace{-1em}
        \caption{\textbf{The impact of random flipping non-salient binarized weights on accuracy in a 1-Bit LLaMA-2-7B.} The x-axis represents the percentage of binarized weights flipped from -1 to 1 or vice versa. As the ratio increases, the accuracy does not decline significantly, indicating redundancy in the 1-bit representation.}
        \label{fig:motivation}
    \vspace{-2em}
    \end{minipage}
    \hfill
    \begin{minipage}{0.48\textwidth}
        \centering
        \includegraphics[width=\linewidth]{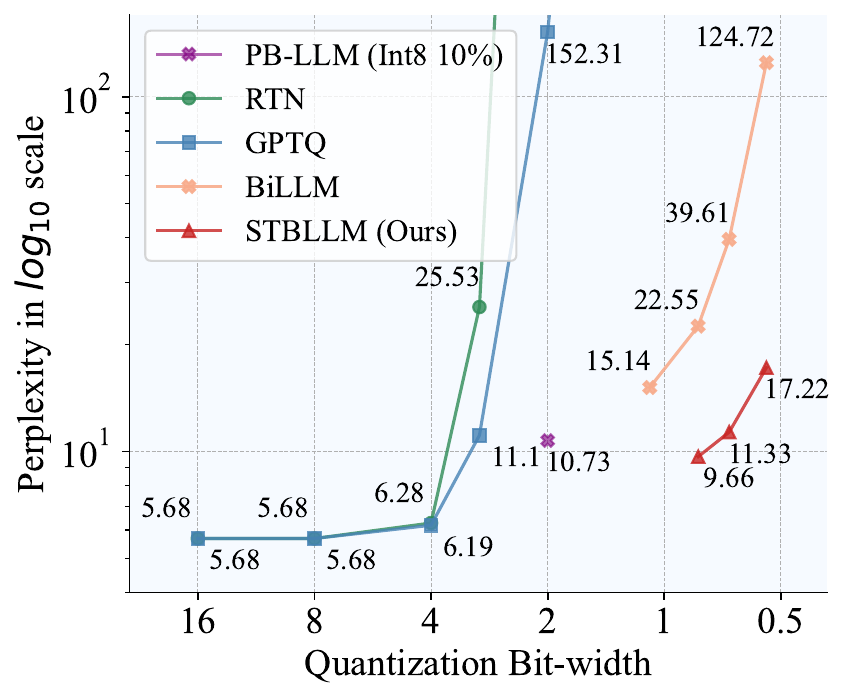}
        \vspace{-1em}
        \caption{\textbf{The perplexity of LLaMA-1-13B on the Wikitext2 under different bit-widths.} RTN and GPTQ~\citep{Frantar2022GPTQAP} show a drastic performance drop at ultra-low bit-widths. Our proposed STBLLM achieves higher performance compared to BiLLM at sub 1-bit widths.}
        \label{fig:bit-vs-ppl-comparison}
    \vspace{-2em}
    \end{minipage}
\end{figure}

Pioneering binarization methods~\citep{ref07_xnor,liu2018bi} present customized binary structures and training paradigms for binarized neural networks (BNNs) in vision tasks. Building upon these foundational approaches, subsequent methods~\citep{Wang2020SparsityInducingBN,Wang2021SubbitNN,Wang2021ExtremelySN,liu2022robust,Li2020BNNPP,Munagala2020STQNetsUN} have advanced the field by integrating sparse kernel techniques~\citep{Wang2023CompactingBN,Wang2021SubbitNN} and pruning methodologies~\citep{Wang2021ExtremelySN,Li2020BNNPP,Munagala2020STQNetsUN}. For LLMs, inspired by the success of 4-bit and 8-bit quantization methods, some studies~\citep{Huang2024BiLLMPT,xu2024onebit, shang2023pb} continue to explore ultra-low-bit or even 1-bit precision. For example, the post-training method PB-LLM~\citep{shang2023pb} partially binarizes LLMs with an optimal scaling factor strategy, preserving a small subset of the higher bit-precision weights. BiLLM~\citep{Huang2024BiLLMPT} proposes a residual approximation strategy to improve 1-bit LLMs. While these methods represent the most aggressive quantization approaches, it is crucial to consider that popular floating-point LLMs already contain model sizes ranging from 7 billion to 140 billion parameters. As a result, 1-bit LLMs still need to be further accelerated and optimized for many resource-constrained devices and real-time scenarios. This naturally raises a key question: \textbf{\emph{Is there any compression method with less than 1-bit weight representation that can further push the quantization of LLMs?}}

For this question, there are two key observations: \textbf{\ding{172} Not all weights contribute equally to the performance of 1-bit  LLMs.} As shown in Figure~\ref{fig:motivation}, performing random weight flipping for non-salient weights results in only a minimal performance drop (For more details, refer to Appendix~\ref{appendix:motivation}). This finding indicates that even in highly quantized 1-bit LLMs, a subset of redundant weights exists that can be compressed without impacting the performance. It suggests the potential for further compression by selectively encoding the most significant weights while discarding or compressing the less important ones. \textbf{\ding{173} Structured sparsity techniques,} such as N:M sparsity methods~\citep{hubara2021accelerated,zhang2022learning,zhou2021learning}, leverage the inherent structure and patterns in the weight distribution, allowing for more efficient compression. These N:M sparsity methods have good hardware-accelerated support in recent LLM pruning models~\citep{frantar2023sparsegpt, Sun2023ASA_wanda,Pruner-Zero}, enabling efficient deployment on NVIDIA Ampere architecture~\citep{nvidia2020}. However, traditional binarization techniques~\citep{ref07_xnor} often treat weights as independent entities, failing to exploit the inherent structure and patterns in the weight matrices. These observations encourage us to explore N:M sparsity tailored specifically for 1-bit LLMs to achieve further speedups and compression gains.

\begin{figure}[t]
    \centering
    \includegraphics[width=1\linewidth]{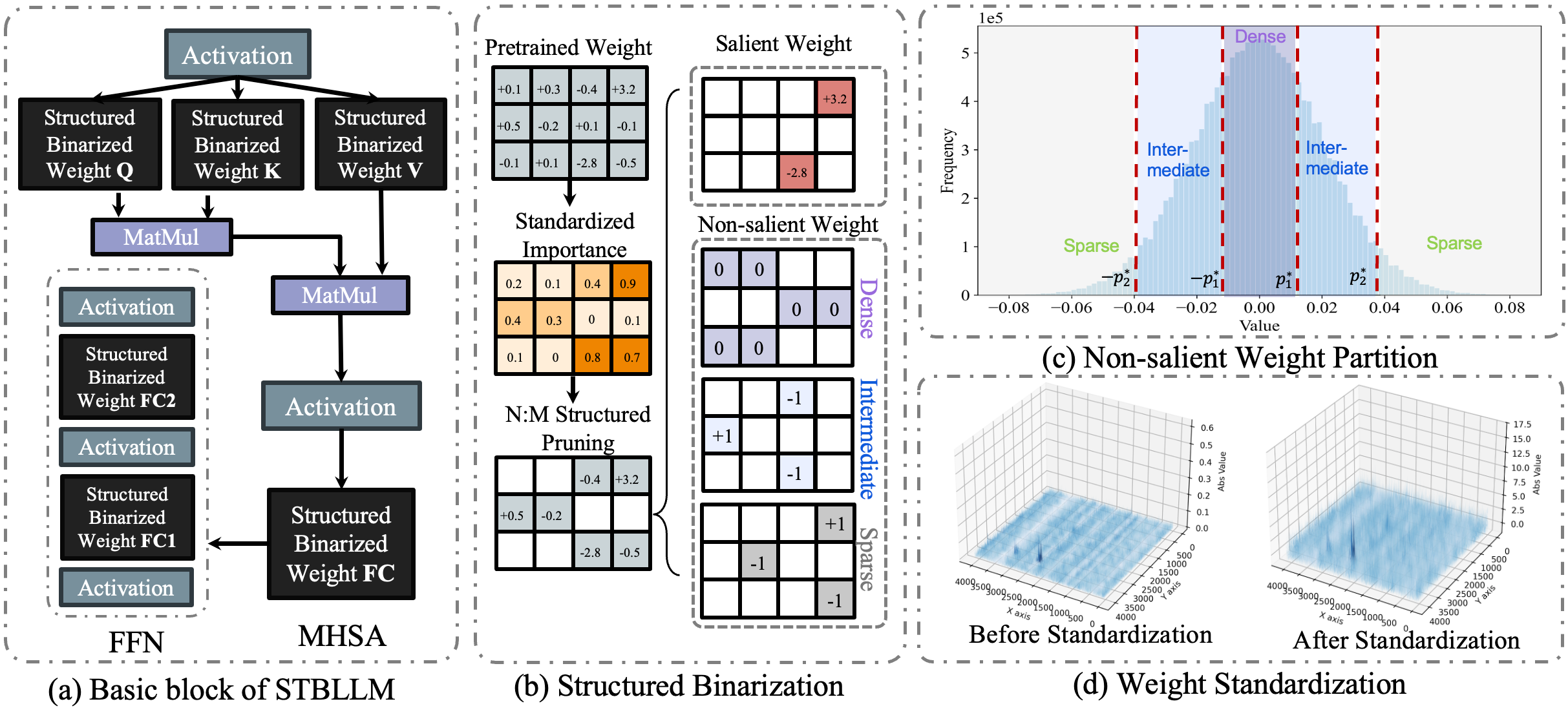}
    \caption{(a) PTQ framework in Structured Binarized LLM (STBLLM). We apply structured binarization to all of the weights. (b) Structured Binarized Weight Computation Procedure. We first perform N:M structure pruning to pre-trained weight (here N=2, M=4), then perform binarization by assign weight to salient and non-salient one. (c) Trisection partition for Symmetric Gaussian Distribution of Non-salient Weight. (d) Illustration of Weight Standardization on LLaMA-2-7B.}
    \label{fig:main_figure}
\vspace{-1.5em}
\end{figure}

Based on these observations, we develop our STBLLM approach,  \underline{ST}ructured \underline{B}inarization for \underline{LLM}s to achieve extreme compression while mitigating performance degradation. Our workflow applies the metric-based sparsity and performs the adaptive N:M  binarization. Specifically, to measure the importance of weights, we introduce a Standardized Importance (SI) metric that addresses the issues of extreme weight values and computationally expensive Hessian-based methods used in prior work. We then propose an adaptive layer-wise structured binarization approach, where different layers of the LLM can be sparsified with varying N:M ratios to balance compression and accuracy. We employ a residual approximation technique~\citep{Huang2024BiLLMPT} for the salient parameters to preserve the critical information. For the non-salient parameters, we utilize a fine-grained grouping strategy based on a trisection search algorithm to find optimal splitting points $p^*$ and apply different quantization schemes to the sparse, intermediate, and dense regions as presented in Figure~\ref{fig:main_figure}(c). By tailoring these structured representations specifically for 1-bit LLMs, we unlock a new avenue for model compression and optimization, enabling more widespread deployment of these powerful LLMs in resource-constrained environments.

To validate the effectiveness of STBLLM, we conduct extensive experiments on various LLMs, including the LLaMA-1/2/3~\citep{touvron2023llama,touvron2023llama2}, OPT~\citep{Zhang2022OPTOP} and Mistral~\citep{Jiang2023Mistral7}. As presented in Figure~\ref{fig:bit-vs-ppl-comparison}, our STBLLM achieves a better trade-off between performance and bit-width. STBLLM with 0.8 bit can achieve lower perplexity than BiLLM with 1.1 bit. STBLLM achieves a perplexity of 31.72 at just 0.55 bits per weight, compared to 688.73 for BiLLM - an over 20$\times$ gain. Even at 65B parameters, our 0.55-bit STBLLM outperforms BiLLM's 0.7-bit and PB-LLM's 1.7-bit versions. STBLLM retains significantly higher accuracy for zero-shot benchmarks than BiLLM under 4:8 and 6:8 structured binarization settings across 13B and 30B LLaMA. For example, on LLaMA-1-30B, our 0.55-bit STBLLM achieves 51.78\% average accuracy versus just 43.72\% for BiLLM. The contribution of our work is as follows:

\begin{itemize}
   \vspace{-0.5em}
    \item We introduce STBLLM, a novel structural binarization framework that compresses Large Language Models (LLMs) to less than 1-bit precision, enabling significant memory and computational savings while preserving model performance.
    \vspace{-0.2em}
    \item STBLLM employs an N:M binary weight kernel approach, where we perform structural binarization of the weights using efficient gradient-free metrics to determine weight importance, channel rearrangement to preserve salient weights, and adaptive layer mixed-structure binarization for better accuracy-efficiency trade-offs.
    \vspace{-0.2em}
    \item We implement a specialized CUDA kernel for structural binarization, leveraging NVIDIA's Ampere GPU sparse tensor cores, achieving a 17.85x speedup over ABQ-LLM’s 2-bit implementation.
   \vspace{-0.2em}
    \item Extensive experiments on various LLMs, including the LLaMA-1/2/3 and OPT, demonstrate STBLLM's superior performance compared to other compressed binarization methods. 
\end{itemize}

\section{Related Work}

\noindent\textbf{Quantization and Binarization.} Quantization reduces full-precision parameters to lower bits, thereby decreasing storage and computation requirements. Recent research has effectively applied Quantization-Aware Training (QAT) and Post-Training Quantization (PTQ) to LLMs. QAT~\citep{liu2023llm_qat, du2024bitdistiller, Chen2024DBLLMAD} incorporates quantization during training, allowing to learn better representations for low-bit weights. However, due to the massive parameters, retraining is too costly and inefficient for LLMs. In contrast, PTQ~\citep{Frantar2022GPTQAP, chee2023quip, lin2023awq, lee2023owq, dettmers2023spqr} can directly quantize pre-trained models without additional training. 
GPTQ~\citep{Frantar2022GPTQAP} and QuIP~\citep{chee2023quip} minimized LLM block quantization errors through second-order error compensation. Other approaches~\citep{lin2023awq,lee2023owq,dettmers2023spqr} focus on prioritizing salient weights to maintain their information representation capacity. Binarization, which constrains quantized parameters to a 1-bit representation, is the most extreme quantization method and has proven effective for vision tasks, such as XNOR-Net~\citep{ref07_xnor} and Bi-Real Net~\citep{liu2018bi}. To further compress binary neural networks, sparse kernel techniques~\citep{Wang2020SparsityInducingBN,Wang2021SubbitNN,Wang2021ExtremelySN,liu2022robust} are introduced to reduce the redundancy in binary neural networks. For LLM binarization, BitNet~\citep{wang2023bitnet} trained 1-bit LLM from scratch. OneBit~\citep{xu2024onebit} employ the QAT paradigm for 1-bit LLM while BiLLM~\citep{Huang2024BiLLMPT} employ the PTQ paradigm with residual approximation technique. In this paper, we further reduce weights below 1-bit by identifying and removing the redundant parameters.

\noindent\textbf{Sparsity Methods for LLM.} Pruning removes less important parameters from a neural network to reduce its size and improve efficiency.
For LLMs, Pruning can be divided to structured pruning~\citep{ma2023llmpruner, ashkboos2024slicegpt,Xia2023ShearedLA, An2023FluctuationbasedAS}, semi-structured pruning~\citep{frantar2023sparsegpt, Sun2023ASA_wanda, zhang2024plugandplay_ria} and unstructured pruning~\citep{frantar2023sparsegpt, Sun2023ASA_wanda, Pruner-Zero}. Structured pruning methods, including LLM-Pruner~\citep{ma2023llmpruner} and Sheared LLaMA~\citep{Xia2023ShearedLA}, aim to simplify LLM by removing specific components such as heads, layers, and dimensions. Although these techniques enhance model efficiency, they often result in significant performance degradation and require extensive retraining to recover lost capabilities. In contrast, unstructured pruning methods~\citep{frantar2023sparsegpt, Sun2023ASA_wanda} remove individual weights based on their significance within the model. However, this approach leads to irregular sparsity patterns that do not effectively leverage hardware acceleration. Semi-structured pruning offers a more balanced approach to model optimization. Methods such as SparseGPT~\citep{frantar2023sparsegpt} and Wanda~\citep{Sun2023ASA_wanda} exemplify this strategy by maintaining regular, hardware-friendly sparsity patterns, such as N:M sparsity. This approach combines the fine-grained control characteristic of unstructured pruning with the operational efficiency associated with structured pruning. 

\noindent\textbf{Synergy of Pruning and Quantization.} The complementary nature of pruning and quantization has been extensively explored in the literature, where pruning reduces the number of parameters in a neural network, and quantization focuses on the precision of those parameters. Deep Compression~\citep{song2016deep_compression} integrates pruning, trained quantization, and Huffman coding into a unified compression pipeline to significantly reduce the storage requirements of deep neural networks (DNNs). Subsequent works have developed in-parallel pruning-quantization methods~\citep{Tung2018CLIPQDN, Yang2019AutomaticNN, Hu2021OPQCD} to optimize compression allocation, such as unstructured pruning sparsity and quantization strategies. For extreme cases like binarization, several approaches~\citep{Munagala2020STQNetsUN, Li2020BNNPP, Wang2021ExtremelySN} combine pruning and compression to achieve high levels of compression and speedup. Specifically, STQ-Nets~\citep{Munagala2020STQNetsUN} extend convolutional neural network (CNN) binarization by incorporating structured pruning, BNN Pruning~\citep{Li2020BNNPP} utilizes weight flipping frequency for further pruning of binary neural networks (BNNs), and BAP~\citep{Wang2021ExtremelySN} introduces binary augmented sparse convolution to attain 98\% sparsity. However, these methods often necessitate a fine-tuning process, which is impractical for LLMs.

\begin{algorithm}[t]
\caption{Framework of STBLLM: Details of each function are shown in Algorithm~\ref{alg2}}
\label{alg1}
\begin{algorithmic}[1]
\Function{StructuredBinaryLLM}{$\mathbf{W}$, $\mathbf{X}$, $\beta$, $\lambda$}
\State \textbf{Input:} $\mathbf{W} \in \mathbb{R}^{n\times m}$ denotes weight matrix; $\mathbf{X} \in \mathbb{R}^{r\times d}$ represents calibration data;
\State $\beta$ denotes block size; $\lambda$ represents hessian regularizer
\State \textbf{Output:} $\mathbf{B}$ - structured binarized weights
\State $\mathbf{H} \gets 2\mathbf{X}\mathbf{X}^\top$ \Comment{$\ell^2$ error hessian matrix}
\State $\mathbf{H}^c \gets \text{Cholesky}({(\mathbf{H} + \lambda \mathbf{I})}^{-1})$
\State $\mathbf{B} \gets 0_{n\times m}$
\For{$b=0, \beta, 2\beta,...,N$}
    \State $\mathbf{W}^{si} \gets \text{Standardized\_Importance}(\mathbf{W}_{:, b:b+\beta})$
    \State $\mathbf{W}^s \gets \text{Semi-Structured}(\mathbf{W}^{si}_{:, b:b+\beta}, \mathbf{W}_{:, b:b+\beta})$
    \State $row_{s}\{\cdot\} \gets \text{Salient}(\mathbf{W}_{:, b:b+\beta}, \mathbf{H}^c)$
    \State $\Tilde{\mathbf{B}}_1 \gets \text{Res\_Approx}(\mathbf{W}_{:, j \in \{ row_{s}\}}^s)$
    \State $p^*_1, p^*_2 \gets \text{NonSalientAwareQuant}(\mathbf{W}_{i,j \notin \{row_{s}\}}^s)$
    \State $\Tilde{\mathbf{B}}_2, \Tilde{\mathbf{B}}_3, \Tilde{\mathbf{B}}_4 \gets \text{Trisection}(\mathbf{W}_{|w_{i,j}|}, p^*_1, p^*_2)$
    \State $\mathbf{B}_{:, b:b+\beta} \gets \Tilde{\mathbf{B}}_1 \cup \Tilde{\mathbf{B}}_2 \cup \Tilde{\mathbf{B}}_3 \cup \Tilde{\mathbf{B}}_4$
    \State $\mathbf{E} \gets (\mathbf{W}_{:, b:b+\beta} - \mathbf{B}_{:, b:b+\beta}) / \mathbf{H}^c_{b:b+\beta, b:b+\beta}$
    \State $\mathbf{W}_{:, b+\beta:} \gets \mathbf{W}_{:, b+\beta:} - \mathbf{E} \cdot \mathbf{H}^c_{b:b+\beta, b+\beta:}$ \Comment{block-wise OBC}
\EndFor
\State \Return $\mathbf{B}$
\EndFunction
\end{algorithmic}
\end{algorithm}

\section{Methodology}

In this section, we introduce our STBLLM framework, as depicted in Figure~\ref{fig:main_figure}. We employ structured binarization for all weights within the Feed-forward Network (FFN) and Multi-head Self-attention (MHSA) modules. Specifically, we introduce the concept of Standardized Importance (SI) to evaluate the saliency of each weight under N:M sparsity constraints (refer to the left part of Figure~\ref{fig:main_figure}(b)). We leverage the Hessian matrix to distinguish between salient and non-salient weights for the binarization process. Salient weights are handled using residual approximation, following the methodology outlined in BiLLM~\citep{Huang2024BiLLMPT}. For non-salient weights, we propose a Non-salient Aware Quantization technique, which further divides these weights into Dense, Intermediate, and Sparse regions (as shown in the right part of Figure~\ref{fig:main_figure}(c)). To optimally partition the non-salient weights into three distinct regions, we utilize a trisection search strategy to determine the appropriate $p^*_1$ and $p^*_2$ values. In the subsequent update step, we apply block-wise error compensation~\citep{frantar2023sparsegpt, Frantar2022GPTQAP} to preserve performance following post-training quantization (PTQ). Algorithm~\ref{alg1} provides a comprehensive overview of the STBLLM process, with detailed implementation steps in Appendix~\ref{appendix:stbllm-impl}.

\subsection{Preliminaries}

\textbf{Binarization.} Binarized compression seeks to quantize floating-point (FP) weights, represented as $\mathbf{W}_{FP}$, into 1-bit values (i.e., $\pm$1). During forward propagation, the sign function is used to binarize the original parameter tensor:
\begin{align}
    \mathbf{B} &:= \alpha \cdot \text{sign}(\mathbf{W}_{FP}), \\
    \operatorname{sign}(w) &:= 
    \begin{cases}
    1& \text{if $x \geq 0$},\\
    -1& \text{others},
    \end{cases}
\end{align} 
where $\mathbf{W}_{FP} \in \mathbb{R}^{n\times m}$ is the 32-bit floating-point weight, and $\mathbf{B}\in \mathbb{R}^{n\times m}$ is the binarized output, and $\alpha := \frac{||\mathbf{W}||_{l_1}}{m}$. The parameter $n$ and $m$ represent the size of the weight matrix. The scaling factor $\alpha \in \mathbb{R}^n$ is applied in a channel-wise manner~\citep{ref07_xnor}.

\textbf{N:M Sparsity.} Inspired by the experiments shown in Figure~\ref{fig:motivation}, we observe the binarized the redundancy in LLMs. By applying the N:M binarization for LLMs, we can achieve an extreme compression ratio of less than 1 bit. Specifically, we introduce an innovative N:M sparsity technique that encodes N consecutive non-zero elements in the weight matrix using a single M-bit representation. Although this approach can accelerate computations, it may lead to performance degradation. To alleviate this problem, we propose several techniques from different perspectives: \ding{172} Importance Measurement. Previous methods~\citep{Frantar2022GPTQAP, Chen2024DBLLMAD, Huang2024BiLLMPT} utilize Hessian-based methods to measure the importance, but these methods can be computationally expensive and may not capture the true importance of parameters in LLMs. \ding{173} Layer-wise Assignment. Previous PTP methods~\citep{frantar2023sparsegpt,zhang2024plugandplay_ria} utilize the uniform sparsity ratio among different layers. However, recently, evidence~\citep{yin2024outlier_owl} shows that not all layers have the same redundant level thus non-uniform sampling can help retain the performance. \ding{174} Hierarchical Quantization. Previous PTQ methods for LLM like AWQ~\citep{lin2023awq}, OWQ~\citep{lee2023owq} and BiLLM~\citep{Huang2024BiLLMPT} split the weights into salient and non-salient parameters using the magnitude of activation or Hessian matrix. They mainly focus on salient weights, as most researchers believe they contribute to the final performance. However, the non-salient parameters also play an essential role in quantization.

\subsection{Standardized Importance Metric} 

Many previous works, such as DB-LLM~\citep{Chen2024DBLLMAD}, SparseGPT~\citep{frantar2023sparsegpt}, GPTQ~\citep{Frantar2022GPTQAP}, and BiLLM~\citep{Huang2024BiLLMPT}, utilize the Hessian metric to measure the importance of weights. However, we observe that extreme values in the weights significantly impact Hessian computation (See Appendix~\ref{appendix:hessian}). To address this issue, we present a Standardized Importance (SI) metric. The computation of SI does not involve the second-order information of the weights, which can be computationally expensive for LLMs. Specifically, we employ standardization to mitigate the issue of extreme values in weights by transforming the weights to have a mean of zero and a standard deviation of one. This process ensures that all weights are on a similar scale, reducing the disproportionate influence of extreme values on the Hessian matrix. For a linear layer with weight $\mathbf{W} \in \mathbb{R}^{n\times m}$, which takes in input activation $\mathbf{X} \in \mathbb{R}^{r\times d}$, where $r$ is the batch size and $d=m$ is the input dimension. We propose to evaluate the importance of each weight by the product of its magnitude and the corresponding input feature norm. The score for the current weight $\mathbf{W}_{i,j}$ is defined as:
\begin{small}
\begin{align}
\mathbf{S}_{i,j}= \sigma(\mu(|\mathbf{W}_{i,j}|)) \cdot ||\mathbf{X}_{:,j}||_2, \quad \sigma(\hat{w})= \frac{w - \mu_{\mathbf{W}}}{\sigma_{\mathbf{W}}},\quad \mu(|\mathbf{W}_{i,j}|)=\frac{|\mathbf{W}_{i,j}|}{\sum_j |\mathbf{W}_{i,j}|} + \frac{|\mathbf{W}_{i,j}|}{\sum_i |\mathbf{W}_{i,j}|},
\end{align}
\end{small}
where $\sigma(\cdot)$ is a normalization function that standardizes the weight magnitude $\mu(|\mathbf{W}_{i,j}|)$ using the mean $\mu_\mathbf{W}$ and standard deviation $\sigma_{\mathbf{W}}$ of all weights in the layer. The weight magnitude $\mu(|\mathbf{W}_{i,j}|)$ is computed as the sum of the L1-normalized magnitude across the input dimension $j$ and the output dimension $i$. The input feature norm $||\mathbf{X}_{:,j}||_2$ is calculated as the L2 norm of the $j$-th column input activation $\mathbf{X}$. By multiplying the standardized weight magnitude $\sigma(\mu(|\mathbf{W}_{i,j}|))$ with the input feature norm $||\mathbf{X}_{:,j}||_2$, the importance score $\mathbf{S}_{i,j}$ takes into account both the significance of the weight itself and the activation level of the associated input feature. To prune the linear layer, we rank all the weights based on their importance scores $\mathbf{S}_{i,j}$ and remove a specified percentage of the weights with the lowest scores. This pruning strategy aims to preserve the most significant weights contributing to the layer's output while eliminating less important weights to reduce the model's size and computational complexity.

\subsection{Adaptive Layer-wise Binarization}

\textbf{N:M Binary Weight Vector.} To achieve compression beyond standard binarization, we propose an $N:M$ sparsity approach, where M binary values are represented by N values (N $<$ M). This allows for further compression while preserving the salient information in the weight tensors. Specifically, we employ the mixed N:8 sparsity configuration following DominoSearch~\citep{sun2021dominosearch}.

\textbf{Layer-wise N:M Assignment.} To achieve better accuracy-efficiency trade-offs, we introduce adaptive layer-wise structured binarization, where different layers of the LLM can be sparsified with different N:M ratios. (For example, with a target ratio of 4:8, layers can have ratios like 3:8, 4:8, and 5:8 while maintaining the overall 4:8 ratio.) This flexibility allows for more aggressive compression in less important layers while preserving higher precision in crucial layers.
The layer-wise N:M ratios are assigned based on the relative importance of each layer, measured by the L2 norm of its weight parameters. Let $\omega_i$ and $\omega_{\text{total}}$ be the L2 norm of layer $i$ and the sum across all layers, respectively. The relative importance $\alpha_i$ of layer $i$ is $\alpha_i = \frac{\omega_i}{\omega_{\text{total}}}$. The N:M ratio for layer $i$ is $\frac{N_i}{M_i} = \alpha_i + (1 - \alpha_i) \cdot R_{\text{target}}$, where $R_{\text{target}}$ is the target overall compression ratio. More important layers have higher N:M ratios (less sparsification), approaching 1:1 for the most important ones. Less important layers have lower N:M ratios, approaching $R_{\text{target}}$ for the least important ones. This ensures the overall compression ratio meets $R_{\text{target}}$.

\subsection{Non-salient Aware Quantization}
Based on the observations that  a small fraction of salient weights is critical to the LLM quantization~\citep{lin2023awq, shao2023omniquant}, we split the weights into the salient and non-salient parts and then apply a higher bit for salient one and lower-bit for non-salient one, as:

\textbf{Salient Part:} In our cases, for salient weight, we apply residual approximation~\citep{Huang2024BiLLMPT}, which is composed of residual approximation weight, as follows: 
\begin{equation}
\begin{cases}
\alpha_o^*, \mathbf{B}_o^* = \arg \min_{\alpha_o, \mathbf{B}_o} \| \mathbf{W} - \alpha_o \mathbf{B}_o \|^2, \\
\alpha_r^*, \mathbf{B}_r^* = \arg \min_{\alpha_r, \mathbf{B}_r} \| (\mathbf{W} - \alpha_o^* \mathbf{B}_o^*) - \alpha_r \mathbf{B}_r \|^2,
\end{cases}
\end{equation}
where $\mathbf{B}_o$ denotes the original binary tensor, and $\mathbf{B}_r$ represent the residual binarized matrix as the compensation. The final approximation of $\mathbf{W}$ is $\mathbf{W} \approx \alpha_{o}^{*} \mathbf{B}_{o}^{*} + \alpha_{r}^{*} \mathbf{B}_{r}^{*}.$ 

\textbf{Non-Salient Part:} 
For the non-salient part (which is also symmetric Gaussian distribution), we find that significant information is retained in the non-salient part. To make the trade-off with bit and performance, we utilize a fine-grained grouping strategy called the Trisection search algorithm (See Algorithm~\ref{alg2}), whose aim is to find the optimal two break-point $p^*_1, p^*_2$. With these two break-points, we can segment the symmetric Gaussian distribution into three groups, which is sparse $R_s[-m,-p^*_2]\cup [p^*_2, m]$, intermediate $R_i[-p^*_2, -p^*_1]\cup [p^*_1, p^*_2]$, and dense region $R_d[-p^*_1, p^*_1]$. Then, we derive the quantization error: 
\begin{align}
\theta^2_{p^*_1, p^*_2} = ||\mathbf{W}_s - \alpha_s \mathbf{B}_s||^2 + ||\mathbf{W}_i - \alpha_i \mathbf{B}_i||^2 + ||\mathbf{W}_d - \alpha_d \mathbf{B}_d||^2, \\ 
\alpha_s = \frac{1}{n_s}||\mathbf{W}_s||_{l1},\ \ \alpha_i = \frac{1}{n_i}||\mathbf{W}_i||_{l1},\ \ \alpha_d = \frac{1}{n_d}||\mathbf{W}_d||_{l1} 
\end{align}
where $\mathbf{W}_s$, $\mathbf{W}_i$, $\mathbf{W}_d$ are the sums of absolute weight values in the sparse, intermediate, and dense regions. $\mathbf{B}_s$, $\mathbf{B}_i$, $\mathbf{B}_d$ are the binarized weights for those regions.
These three regions are binarized separately. This method introduces an additional 2 bits for group identification, which constitutes a minor portion of the overall bit count, while the majority of computing parameters remain at 1 bit.

\textbf{Average Bits.} In STBLLM, we introduce extra bits while pruning the redundant or less important weights. The overhead of weight parameters is $N_\text{param} = 2 \times r_\text{salient} + 1 \times (1 - r_\text{salient})$. The additional hardware overhead is $N_\text{storing} = 2 + \frac{1}{b_\text{size}}$, where $r_{salient}$ denotes the proportion of salient weights and $b_\text{size}$ denotes the block size in OBC compensation, with 2 bits allocated for marking the division of non-salient weights. Under N:M binarization settings, where N and M are positive integers with N < M, we prune the model weights by retaining only a fraction (N/M) of the original weights. Consequently, the number of parameters in the pruned STBLLM model is $N_\text{stbllm}=N_\text{param}\times \frac{N}{M}$. This N:M binarization method allows for a significant reduction in model size.

\begin{table}[t]
\centering
\caption{Average bit results from structural searching and residual binarization of OPT, LLaMA-1, and LLaMA-2 families. *OPT-66B, LLaMA-1-65B and LLaMA-2-70B.}\label{tab:average_bits}
\vspace{-0.3em}
\resizebox{\textwidth}{!}{
\begin{tabular}{lcccccccccccccccc}
\toprule
\textbf{Model} & \multicolumn{4}{c}{\textbf{BiLLM}} & \multicolumn{4}{c}{\textbf{BiLLM-4:8}} & \multicolumn{4}{c}{\textbf{BiLLM-5:8}} & \multicolumn{4}{c}{\textbf{BiLLM-6:8}} \\
\cmidrule(lr){2-5} \cmidrule(lr){6-9} \cmidrule(lr){10-13} \cmidrule(lr){14-17}
 & 7B & 13B & 30B & 65-70B* & 7B & 13B & 30B & 65-70B* & 7B & 13B & 30B & 65-70B* & 7B & 13B & 30B & 65-70B* \\
\midrule
OPT & 1.10 & 1.12 & 1.12 & 1.13 & 0.55 & 0.56 & 0.56 & 0.56 & 0.69 & 0.70 & 0.70 & 0.71 & 0.83 & 0.84 & 0.84 & 0.85 \\
LLaMA-1 & 1.09 & 1.09 & 1.10 & 1.10 & 0.54 & 0.54 & 0.55 & 0.55 & 0.68 & 0.68 & 0.69 & 0.69 & 0.82 & 0.82 & 0.83 & 0.83 \\
LLaMA-2 & 1.07 & 1.08 & N/A & 1.09 & 0.53 & 0.54 & N/A & 0.54 & 0.67 & 0.67 & N/A & 0.68 & 0.80 & 0.81 & N/A & 0.82 \\
\bottomrule
\end{tabular}
}
\vspace{-0.5em}
\end{table}

\begin{table*}[t]
\renewcommand{\arraystretch}{1.0}
\small
\centering
\setlength{\tabcolsep}{1.50mm}
\caption{Perplexity comparison of PB-LLM and BiLLM on the LLaMA model family. The columns represent the perplexity results on the Wikitext2 for different model sizes. The average bit-width for each model is provided in the table. For more precise bit-width results, please refer to Table~\ref{tab:average_bits}. }\label{tab:llama_series}
\vspace{-0.4em}
\resizebox{0.9\textwidth}{!}{
\begin{tabular}{lccrrrrrrr}
    \toprule
    \multicolumn{3}{c}{\textbf{Settings}} & \multicolumn{4}{c}{\textbf{ LLaMA-1}}  & \multicolumn{2}{c}{\textbf{ LLaMA-2}} & \multicolumn{1}{c}{\textbf{ LLaMA-3}} \\
    \midrule
   Method & Block Size & \multicolumn{1}{p{4.19em}}{W-Bits} & 7B    & 13B   & 30B   & 65B   & 7B    & 13B   & 8B \\
    \midrule
    FullPrecision & -     & 16    & 5.68  & 5.09  & 4.1   & 3.53  & 5.47  & 4.88  & 6.10 \\
    RTN           &   -  & 1     & 1.7e5 & 1.4e6 & 1.5e4 & 6.5e4 & 1.6e5 & 4.8e4 & 2.7e6 \\
    GPTQ     & 128     & 1 & 2.7e5       & 1.1e5 & 6.7e4       & 2.5e4       & 1.2e5       & 9.4e3       & 5.7e4      \\ 
    PB-LLM & 128   & 1.7   & 102.36 & 36.6  & 33.67 & 12.53 & 69.2  & 151.09 & 41.80 \\
    BiLLM & 128   & 1.09  & 35.04 & 15.14 & 10.52 & 8.49  & 32.48 & 16.77 & 28.30 \\
    \midrule
    BiLLM & 128   & 0.80 (6:8)   & 80.36 & 22.55 & 13.22 & 9.09   & 50.25 & 27.28 & 94.15 \\
    BiLLM & 128   & 0.70 (5:8)   & 126.99 & 39.61 & 18.69 & 11.57   & 87.84 & 58.14 & 161.48 \\
    BiLLM & 128   & 0.55 (4:8)   & 688.73 & 124.72 & 37.96 & 29.22   & 263.61 & 124.78 & 663.91 \\ \midrule
    STBLLM & 128   & 0.80 (6:8)   & 15.03 & 9.66  & 7.56  & 6.43  & 13.06 & 11.67 & 33.44 \\
    STBLLM & 128   & 0.70 (5:8)   & 19.48 & 11.33 & 9.19  & 7.91  & 18.74 & 13.26 & 49.12 \\
    STBLLM & 128   & 0.55 (4:8)   & 31.72 & 17.22 & 13.43 & 11.07 & 27.93 & 20.57 & 253.76 \\
    \bottomrule
    \end{tabular}}
\vspace{-0.3em}
\end{table*}

\begin{table}[t]
\centering
\caption{Perplexity results on Wikitext2 datasets of OPT and Mistral models with BiLLM and STBLLM. For more precise bit-width results, please refer to Table~\ref{tab:average_bits}. }
\vspace{-0.5em}
\resizebox{0.65\textwidth}{!}{
  \setlength{\tabcolsep}{5.5pt}
    \begin{tabular}{rcrrrcr}
    \toprule
        \multicolumn{2}{c}{\textbf{Settings}}      & \multicolumn{4}{c}{\textbf{OPT}}       & \multicolumn{1}{c}{\textbf{Mistral}} \\
    \midrule
    Method & \multicolumn{1}{c}{W-Bits} & 1.3B  & 2.7B  & 6.7B  & \multicolumn{1}{r}{30B} & 7B  \\
    \midrule
    BiLLM & 0.80 (6:8)   & 51.62 & 23.03 & 15.82 & 15.82 & 72.29 \\
    BiLLM & 0.70 (5:8)   & 69.15 & 30.62 & 20.58 & 20.58 & 82.84 \\
    BiLLM & 0.55 (4:8)   & 106.99 & 55.28 & 79.68 & 79.68 & 189.73 \\ \midrule
    STBLLM & 0.80 (6:8)   & 29.84 & 17.02 & 12.79 & 12.80 & 27.31 \\
    STBLLM & 0.70 (5:8)   & 33.01 & 20.82 & 14.38 & 14.38 & 25.64 \\
    STBLLM & 0.55 (4:8)   & 45.11 & 30.34 & 18.80 & 18.80 & 70.14 \\
    \bottomrule
    \end{tabular}
}
\label{table:performance_opt_mistral}
\vspace{-1.2em}
\end{table}

\section{Experiments}

\subsection{Implementation Details}

\textbf{Experimental Setup.} Our STBLLM utilizes PyTorch~\citep{paszke2019pytorch} and Huggingface~\citep{wolf2019huggingface} libraries. Most LLMs except 65B can be evaluated on a single NVIDIA A800 GPU. For the LLaMA-1-65B model, we employ four NVIDIA A800 GPUs for evaluation. 
It takes 1.8 hours for the post-training process of 7B models on an RTX 4090 GPU and 2.8 hours for 13B models on an A6000 GPU. Following BiLLM~\citep{Huang2024BiLLMPT}, our proposed STBLLM also eliminates the need for fine-tuning, offering an efficient post-training quantization framework. 

\textbf{Datasets and Models.} We measure the perplexity for language generation tasks on Wikitext2~\citep{merity2016pointer_wikitext2}, C4~\citep{2019t5_c4dataset} and PTB~\citep{marcus-etal-1993-building-ptb}, and accuracy for the zero-shot tasks including Winogrande~\citep{sakaguchi2021winogrande}, OBQA~\citep{OpenBookQA2018}, Hellaswag~\citep{zellers2019hellaswag}, BoolQ~\citep{clark2019boolq}, ARC~\citep{clark2018think_arc} and RTE~\citep{Chakrabarty2021FigurativeLI_RTE}. We conduct  experiments on LLaMA-1/2/3~\citep{touvron2023llama,touvron2023llama2}, OPT~\citep{Zhang2022OPTOP}, and Mistral~\citep{Jiang2023Mistral7}. For perplexity evaluation in Table~\ref{tab:llama_series} and ~\ref{table:performance_opt_mistral}, we employ the C4 dataset as the calibration dataset and report the perplexity on Wikitext2.

\begin{table*}[t]
  \centering
    \caption{
  Accuracies (\%) for 7 zero-shot tasks from structured binarized LLaMA-1-13B, LLaMA-2-13B, and LLaMA-1-30B with BiLLM and STBLLM. We compare the performance under the same N:M setting to achieve sub-1-bit quantization.
  }
\vspace{-0.5em}
\resizebox{1.0\textwidth}{!}{
  \setlength{\tabcolsep}{5.5pt}
    \begin{tabular}{llcccccccc}
    \toprule
    \textbf{Models} & \textbf{Method} & \textbf{Winogrande} & \textbf{OBQA}  & \textbf{Hellaswag} & \textbf{Boolq} & \textbf{ARC-e}  & \textbf{ARC-c}  & \textbf{RTE}   & \textbf{Mean} \\
    \midrule
    \multirow{5}[2]{*}{LLaMA-1-13B} & FullPrecision & 72.77  & 33.20  & 59.94  & 77.89  & 77.40  & 46.50  & 70.40  & 62.59  \\
          & BiLLM(6:8) & 58.80  & 30.60  & 46.25  & 62.96  & 49.96  & 23.97  & 53.42  & 46.57  \\
          & BiLLM(4:8) & 52.09  & 28.00  & 30.82  & 61.25  & 32.66  & 21.25  & 53.07  & 39.88  \\
          & STBLLM(6:8) & 65.98  & 36.20  & 63.67  & 65.38  & 68.86  & 34.04  & 56.68  & 55.83  \\
          & STBLLM(4:8) & 63.06  & 34.80  & 52.65  & 62.48  & 56.90  & 28.33  & 52.71  & 50.13  \\
    \midrule
    \multirow{5}[2]{*}{LLaMA-2-13B} & FullPrecision & 72.22  & 35.20  & 60.06  & 80.52  & 79.42  & 48.46  & 65.34  & 63.03  \\
          & BiLLM(6:8) & 56.43  & 30.60  & 35.53  & 62.48  & 41.29  & 24.74  & 53.43  & 43.50  \\
          & BiLLM(4:8) & 50.59  & 24.00  & 28.96  & 62.08  & 30.51  & 22.35  & 53.07  & 38.79  \\
          & STBLLM(6:8) & 63.93  & 37.00  & 57.76  & 71.53  & 60.56  & 31.99  & 54.15  & 53.85  \\
          & STBLLM(4:8) & 55.88  & 29.40  & 44.03  & 64.31  & 48.86  & 26.54  & 52.71  & 45.96  \\
    \midrule
    \multirow{5}[2]{*}{LLaMA-1-30B} & FullPrecision & 75.69  & 36.00  & 63.35  & 82.69  & 80.30  & 52.82  & 66.79  & 65.38  \\
          & BiLLM(6:8) & 66.54  & 36.40  & 58.18  & 66.15  & 62.37  & 31.91  & 46.93  & 50.32  \\
          & BiLLM(4:8) &54.93 & 29.40 & 38.85 & 62.17 & 43.6 & 24.74 & 52.35 & 43.72 \\
          & STBLLM(6:8) & 71.59 & 41.00 & 69.85 & 77.37 & 71.55 & 41.3 & 48.01 & 60.10 \\
          & STBLLM(4:8) & 64.01 & 34.60 & 56.46 & 63.06 & 60.86 & 31.48 & 51.99 & 51.78 \\
    \bottomrule
    \end{tabular}%
}
\vspace{-1.4em}
\label{tab:zero_shot_main_detail}
\end{table*}

\textbf{Baseline.} Our primary baseline is BiLLM~\citep{Huang2024BiLLMPT}, which is a 1-bit PTQ framework for LLMs. We perform an N:M sparse pattern on pre-trained LLMs and then conduct the same procedure as BiLLM to report the results that are less than 1 bit (e.g. 0.8, 0.7, 0.55 bits). We conduct the  N:M sparsity using Wanda~\citep{Sun2023ASA_wanda} as the baseline, a gradient-free post-training pruning method. We compare the results of STBLLM with BiLLM under the same N:M settings. For more information on average bits under N:M settings, please refer to Table~\ref{tab:average_bits}. Previous low-bit methods like PB-LLM~\citep{shang2023pb}, GTPQ~\citep{Frantar2022GPTQAP} and vanilla RTN are also selected for comparison. 

\subsection{Main Results}

\textbf{Comparison with PTQ methods.} 
We comprehensively compare the performance of different LLaMA families across various model sizes (7B-65B). For a fair comparison, we set the same block size to 128. As presented in Table~\ref{tab:llama_series}, the model under RTN and GPTQ fails to retain the performance at 1-bit. PB-LLM has shown a satisfactory perplexity under 1.7 bit but deteriorates performance compared with BiLLM under 1.09 bit. To further compare the performance at sub-1-bit, we apply the same N:M setting to BiLLM and our proposed STBLLM. As shown in Figure~\ref{fig:bit-vs-ppl-comparison}, our proposed STBLLM achieves a better trade-off between bit-widths and perplexity across model sizes from 7B to 65B.
STBLLM surpasses BiLLM by a large margin ($688.73\rightarrow 31.72$) on LLaMA-1-7B, especially on the most extreme compression case, 4:8 structured binarization, which means setting half of the parameter to zero. 
It is also noteworthy that when the parameter size reaches 65B, our STBLLM, at 0.55 bit, achieves a perplexity of 11.07, surpassing that of PB-LLM (12.53) at 1.7 bit and that of BiLLM (11.57) at 0.7 bit. 
To our knowledge, our STBLLM is the first work that breaks the 1-bit barriers by further reducing the redundant weights in an N:M pattern. 
Moreover, we conduct further experiments on the OPT family from 1.3B to 30B and Mistral-7B at sub-1-bit PTQ settings. From Table~\ref{table:performance_opt_mistral}, we observe the same trend as in LLaMA. Our proposed STBLLM performs significantly better than BiLLM across all models and all N:M structured binarization settings.

\begin{minipage}[t]{0.52\textwidth}
    \centering
    \captionof{table}{Ablation for pruning metric.}
    \resizebox{\textwidth}{!}{
        \begin{tabular}{lcccc}
        \toprule
        \textbf{Model} & \textbf{Magnitude} & \textbf{Wanda} & \textbf{SparseGPT} & \textbf{Ours (SI)} \\  \midrule
        LLaMA-1-7B & 4797.41 & 207.32 & 32.82 & 31.72 \\ 
        LLaMA-2-7B & 2287.24 & 97.54 & 31.55 & 27.93 \\ 
        \bottomrule
        \end{tabular}
    }
    \label{tab:comparison_pruning_methods}
\end{minipage}%
\hfill
\begin{minipage}[t]{0.48\textwidth}
    \centering
    \captionof{table}{Ablation study for allocation strategy.}
    \resizebox{0.8\textwidth}{!}{
        \begin{tabular}{cccc}
        \toprule
          \textbf{Models}    & \textbf{Uniform} & \textbf{Sin-shape} & \textbf{Ours} \\
        \midrule
        LLaMA-1-7B & 80.36 & 67.78 & 15.03 \\
        LLaMA-2-7B & 50.25 & 33.61 & 13.06 \\
        \bottomrule
        \end{tabular}
    }
    \label{tab:allocation}
\end{minipage}

\textbf{Zero-Shot Performance.}
To conduct a more comprehensive evaluation of binary LLMs, we extend our experiments to 7 zero-shot datasets on LLaMA-1-13B, LLaMA-2-13B, and LLaMA-1-30B, each tested with FullPrecision, BiLLM(6:8), BiLLM(4:8), STBLLM(6:8), and STBLLM(4:8) methods. We mainly focus on the performance of these models under the sub-1-bit setting. Specifically, we compare the BiLLM and our STBLLM under 4:8 and 6:8 structured binarization settings. As illustrated in Table~\ref{tab:zero_shot_main_detail}, we find that the performance drop in reduced precision methods is more pronounced in BiLLM methods compared to STBLLM methods, indicating that STBLLM methods are more robust alternatives when memory resources are constrained.

\subsection{Enhancing Inference Efficiency on Hardware.} 

\begin{figure}
    \centering
    \begin{subfigure}[b]{0.48\linewidth}
        \centering
        \includegraphics[width=\linewidth]{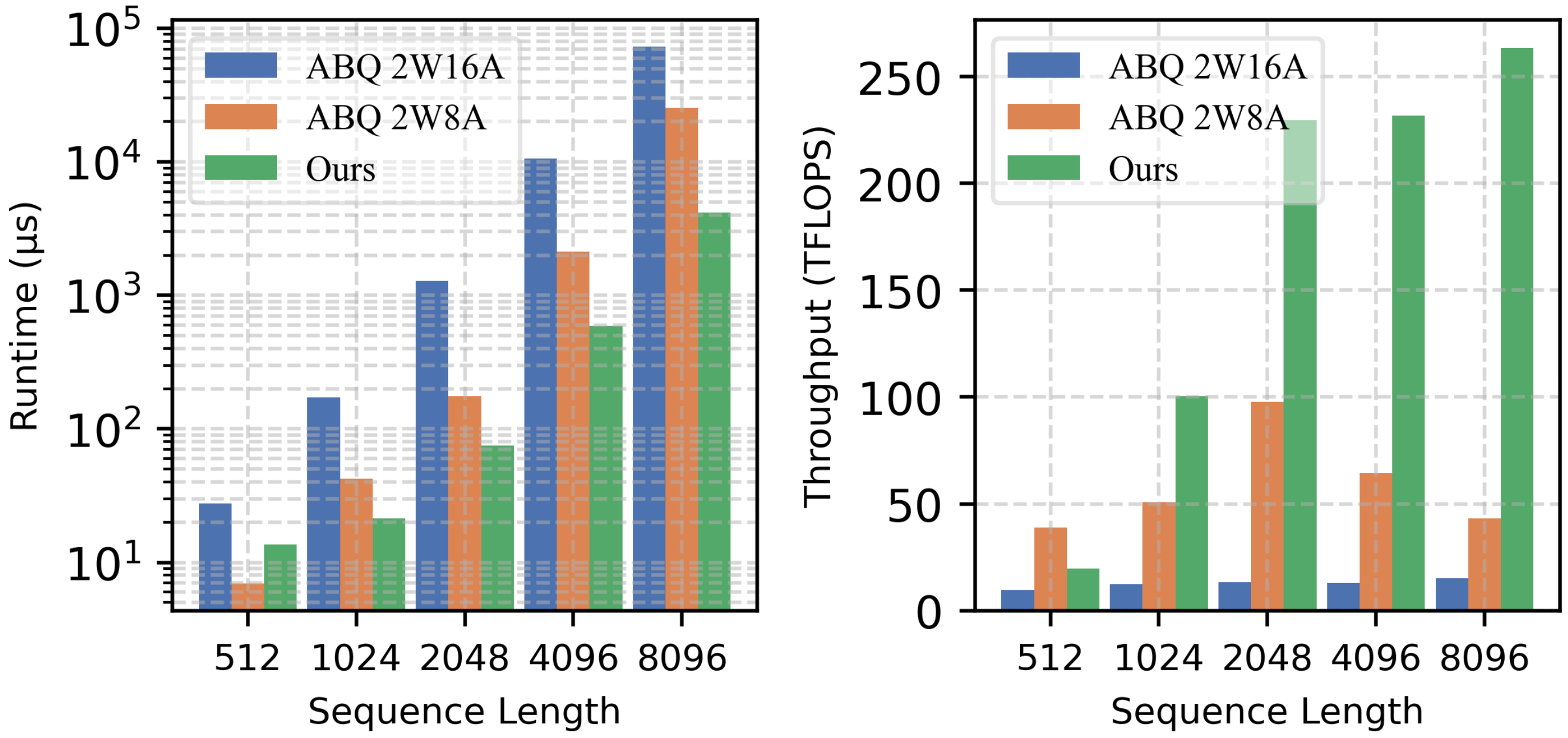}
        \caption{Runtime and throughput comparison.}
        \vspace{-0.5em}
        \label{fig:subfig1}
    \end{subfigure}
    \hfill
    \begin{subfigure}[b]{0.48\linewidth}
        \centering
        \includegraphics[width=\linewidth]{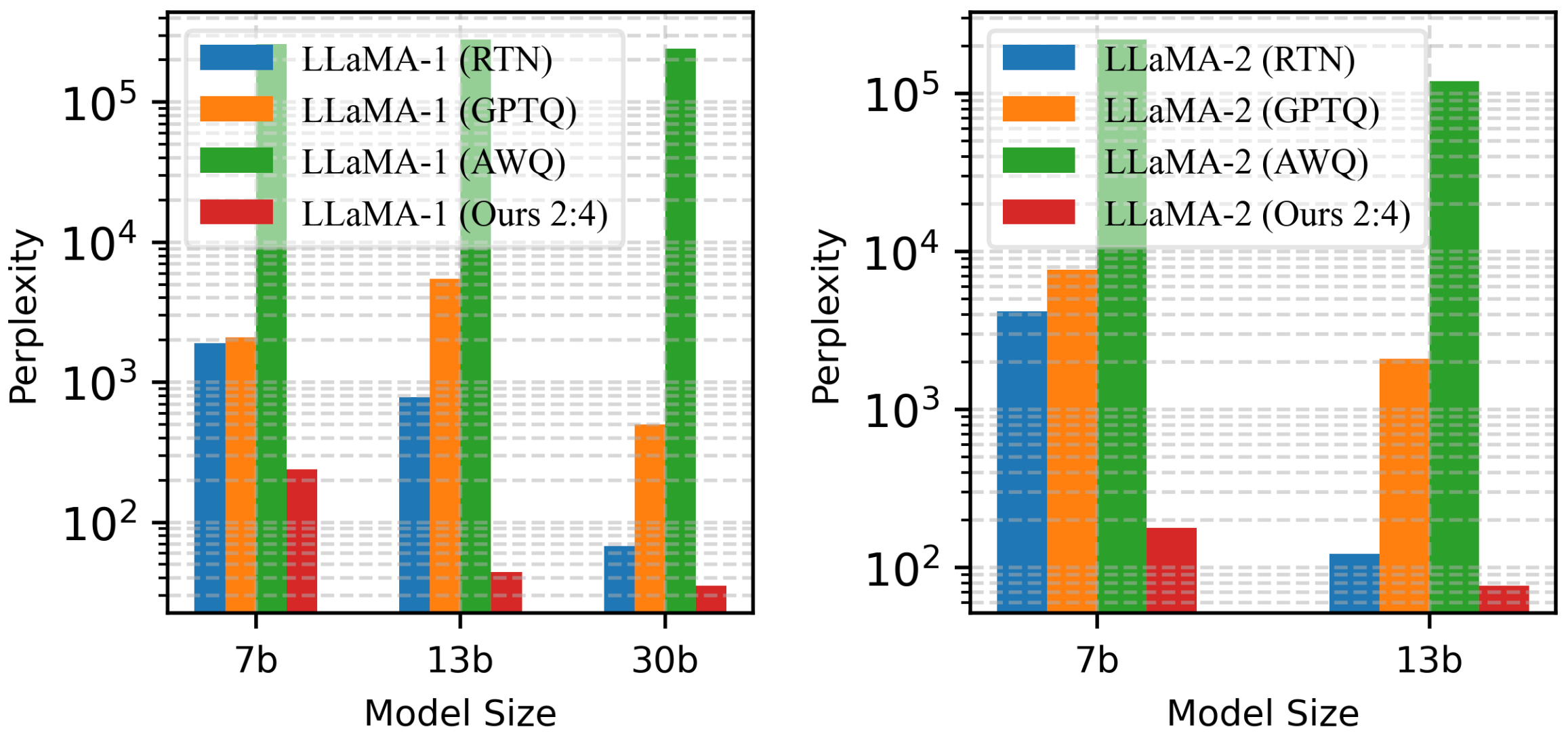}
        \caption{Perplexity comparison.}
        \vspace{-0.5em}
        \label{fig:subfig2}
    \end{subfigure}
    \caption{(a) Runtime and throughput comparison across sequence lengths for ours and ABQ-LLM. (b) Perplexity comparison across model sizes under 2:4 setting for LLaMA-1/2.}
    \label{fig:hardware_speedup}
    \vspace{-1.5em}
\end{figure}

\vspace{-0.5em}
We present specialized CUDA kernels designed to support 1-bit 2:4 sparsification. As illustrated in Figure~\ref{fig:hardware_speedup}(a), we utilize a 2-bit implementation on recent RTX4090 GPU from ABQ-LLM~\citep{zeng2024abq} as the baseline (W2A16 and W2A8) and compare it with our highly-optimized 2:4 1-bit implementation. We provide a comparative analysis of runtime and throughput across various sequence lengths, demonstrating the significant gains in computational efficiency and reduced memory footprint. 
Specifically, for typical sequence lengths of 4096 and 8192, our implementation achieves up to 17.85 times speedup compared to ABQ-LLM's 2-bit implementation. At a sequence length of 8192, our kernel reaches 263.45 TFLOPS, which is 79.74\% of the RTX4090's 2:4 sparse tensor core peak performance. Notably, the speedup becomes more pronounced as sequence length increases. Furthermore, as illustrated in Figure~\ref{fig:hardware_speedup} (b), our method yields lower perplexity for LLaMA-1/2 models. This indicates superior model performance and accuracy compared to 2-bit round-to-nearest (RTN), GPTQ, and AWQ. Refer to Appendix~\ref{appendix:hardware} for the memory comparison and implementation details.

\subsection{Ablation Studies}
\vspace{-0.5em}

\textbf{Ablation for Metric.} Table~\ref{tab:comparison_pruning_methods} shows the impact of post-training pruning metrics (Magnitude, Wanda~\citep{Sun2023ASA_wanda}, SparseGPT~\citep{frantar2023sparsegpt} and our SI) on STBLLM regarding LLaMA-1-7B and LLaMA-2-7B. During PTP, we employ the C4 dataset as the calibration dataset and report the perplexity on the Wikitext2 dataset. SparseGPT requires second-order information, which involves a massive computation burden. Similar to Wanda, our SI does not require gradient or second-order information. Our method achieves better performance among these metrics. 

\textbf{Ablation for Quantization Strategy.} We conduct an ablation study on different quantization strategies. Comparing the perplexity of our Non-salient Aware Quantization (dubbed as Non-salient) and Bell-shaped Distribution Splitting (dubbed as Bell-shaped) in BiLLM~\citep{Huang2024BiLLMPT} on both LLaMA-1-7B and LLaMA-2-7B, as shown in Table~\ref{tab:mask_comparison}. The perplexity of Non-salient changes a lot when moving from LLaMA-1-7B to LLaMA-2-7B, while our Non-salient exhibits nearly identical perplexity in both models, significantly lower than that of Bell-shaped.

\textbf{Ablation for Allocation Strategy.} Table~\ref{tab:allocation} presents an ablation study on different allocation strategies. We compare our method with Uniform and Sin-shaped allocation strategies. The Sin-shaped strategy assigns layer-wise sparsity following a sine wave pattern, where the initial layers have lower sparsity and the latter have higher sparsity. The performance of Uniform and Sin-shaped strategies varies significantly across different models. In contrast, our strategy consistently achieves nearly identical performance across both models, outperforming the other two allocation strategies.

\textbf{Ablation for Group Size.} Table~\ref{tab:group_size} presents the results of our ablation study on the group size configuration. We evaluate the perplexity of LLaMA-1-7B and LLaMA-2-7B with group sizes of 64, 128, 256, and 512. Generally, as the group size increases, performance improves. However, this also results in higher computational and storage demands. We choose a group size of 128 to balance performance and resource consumption.

\begin{table}[t]
    \centering
    \caption{Comparison of Magnitude, Wanda, SparseGPT, and SI across different datasets.}
    \vspace{-0.4em}
      \resizebox{1.0\textwidth}{!}{
    \begin{tabular}{lcccc|lcccc}
    \toprule
    Models & \multicolumn{3}{c}{LLaMA-1-7B} &       & Models & \multicolumn{4}{c}{LLaMA-2-7B} \\
    \midrule
    Dataset & Magnitude & Wanda & SparseGPT & Ours(SI)  & Dataset & Magnitude & Wanda & SparseGPT & Ours(SI) \\
    \midrule
    PTB   & 11608.88 & 306.57 & 61.53 & 68.48 & PTB   & 45564.36 & 2027.33 & 236.03 & 690.76 \\
    C4    & 1545.34 & 153.29 & 33.06 & 36.04 & C4    & 1034.84 & 86.45 & 30.53 & 30.81 \\
    Wikitext2 & 4797.42 & 207.32 & 32.82 & 31.72 & Wikitext2 & 2287.25 & 97.54 & 31.56 & 27.93 \\
    \bottomrule
    \end{tabular}%
}
\label{tab:comparison_ablation_metric_dataset}
\end{table}
\vspace{-0.4em}

\begin{minipage}[t]{0.45\textwidth}
\centering
\captionof{table}{Ablation for quantization strategy.}
\vspace{-0.2em}
\resizebox{0.9\textwidth}{!}{
    \begin{tabular}{ccc}
        \toprule
      \textbf{Models}  & \textbf{Bell-shaped} & \textbf{Non-salient} \\
        \midrule
        LLaMA-1-7B & 80.35 & 15.03 \\
        LLaMA-2-7B & 50.25 & 13.06 \\
        \bottomrule
    \end{tabular}
}
\label{tab:mask_comparison}
\end{minipage}
\hfill 
\begin{minipage}[t]{0.55\textwidth}
\makeatletter\def\@captype{table}
\centering
\captionof{table}{Ablation for group size.}\label{tab:group_size}
\vspace{-0.2em}
\resizebox{.9\textwidth}{!}{
    \begin{tabular}{lcccccc} 
    \toprule 
    \textbf{Model} & \textbf{{64}} & \textbf{{128}} & \textbf{{256}} & \textbf{{512}} & \textbf{{1024}} \\
    \midrule 
    {LLaMA-1-7B} & 29.58 & 31.72 & 33.97 & 41.29 & 146.46 \\
    {LLaMA-2-7B} & 27.12 & 27.93 & 50.62 & 54.68 & 507.44 \\
    \bottomrule 
    \end{tabular} 
}
\label{tab:comparison_allocation_strategy}
\end{minipage}

\section{Conclusion}

In this paper, we introduce STBLLM, a structured Binary LLM PTQ framework designed for sub-1-bit quantization. We address redundancy in binarized LLMs, highlighting the potential for further compression. Specifically, we present a Standardized Importance (SI) metric for N:M structured pruning. Then, we use the Hessian matrix to partition weights into salient and non-salient categories. We propose Non-salient Aware Quantization for non-salient weights, identifying optimal splitting points to create sparse, intermediate, and dense regions, each with tailored binarization. Finally, we design a specialized CUDA kernel with a sparse tensor core to achieve significant speedup. We validate the performance of STBLLM across LLaMA-1/2/3, OPT, and Mistral, demonstrating that STBLLM achieves a superior trade-off at sub-1-bit settings. By achieving LLM performance under 1 bit, STBLLM highlights the potential of extreme LLM compression. \textbf{Limitation:} STBLLM does not support Mixture of Experts (MoE) or Mamba-based language models.

\newpage 

\bibliography{iclr2025_conference}
\bibliographystyle{iclr2025_conference}

\newpage 

\appendix

\section*{Appendix Overview}

\begin{itemize}
    \item Section~\ref{appendix:stbllm-impl}: STBLLM Implementation. 
    \item Section~\ref{appendix:motivation}: Details of Motivation Experiments. 
    \item Section~\ref{appendix:hardware}: Details of Hardware Accelerations. 
    \item Section~\ref{appendix:ablation}: More Experimental Results. 
    \item Section~\ref{appendix:hessian}: Impact of Extreme Weight on the Hessian Matrix.
\end{itemize}

\section{STBLLM Implementation}
\label{appendix:stbllm-impl}

Following BiLLM~\citep{Huang2024BiLLMPT}, STBLLM does not change the operations on salient weights. Instead, STBLLM mainly focuses on the non-salient weight. We present \textbf{NonSalientAwareQuant} and \textbf{Trisection} function in Algorithm~\ref{alg2}. 

For \textbf{NonSalientAwareQuant} function, it aims to find two optimal break-points to partition the symmetric Gaussian distribution of non-salient weight. A naive approach for searching the break-point is using two nested loops, whose complexity is $O(N^2)$, where $N$ denotes the length of the search space. To reduce the complexity to $O(N)$, we propose to utilize $p_2=\sigma \times p_1$ to locate the $p_2$. It is natural to assume that $p_2 > p_1$ and we have $\sigma > 1$. In practice, we set the $\sigma=2$ and it works well. 

For \textbf{Trisection} function, it aims to partition the symmetric Gaussian distribution presented in Figure~\ref{fig:main_figure}(c) into three parts, which are Sparse, Intermediate, and Dense region. These three parts have no intersection and by uniting them together, we have all of the non-salient structured binarized weight. 

\begin{algorithm*}[h]
\small 
\begin{multicols}{2}
\caption{STBLLM}
\label{alg2}
func $\operatorname{Salient}$ $(\mathbf{W}, \mathbf{H^c})$
\begin{algorithmic}[1]
\Function{Salient}{$\mathbf{W}, \mathbf{H^c}$}
    \State $\mathbf{S} \gets \mathbf{W}^2 / [\mathbf{H}^c_{b:b+\beta; b:b+\beta}]^2$ \Comment{Salient matrix}
    \State $row_s \gets \operatorname{topk}(\operatorname{sum}(\operatorname{abs}(\mathbf{S})), \text{dim}=0)$
    \State $e \gets \infty$ \Comment{Searching error}
    \State $n^* \gets 0$ \Comment{Optimal number of salient columns}
    \For{$i = 1$ to $\operatorname{len}(row_s)$}
        \State $\mathbf{B}_1 \gets \operatorname{binary}(\mathbf{W}_{:,j}, j \in row_s[:i])$
        \State $\mathbf{B}_2 \gets \operatorname{binary}(\mathbf{W}_{:,j}, j \notin row_s[:i])$
        \If{$\|\mathbf{W} - (\mathbf{B}_1 \cup \mathbf{B}_2)\|^2 < e$}
            \State $e \gets \|\mathbf{W} - (\mathbf{B}_1 \cup \mathbf{B}_2)\|^2$
            \State $n^* \gets i$
        \EndIf
    \EndFor
    \State \Return $row_s[:n^*]$
\EndFunction
\end{algorithmic}

\hspace{0.04cm}

\begin{algorithmic}[1]
\Function{Binary}{$\mathbf{W}$}
    \State $\alpha \gets \frac{\|\mathbf{W}\|_{\ell1}}{m}$
    \State $\mathbf{B} \gets \alpha \cdot \operatorname{sign}(\mathbf{W})$
    \State \Return $\mathbf{B}$
\EndFunction
\end{algorithmic}

\hspace{0.05cm}

\begin{algorithmic}[1]
\Function{Res\_Approx}{$\mathbf{W}$}
    \State $\mathbf{B}_1 \gets \Call{Binary}{\mathbf{W}}$
    \State $\mathbf{R} \gets \mathbf{W} - \mathbf{B}_1$
    \State $\mathbf{B}_2 \gets \Call{Binary}{\mathbf{R}}$
    \State $\mathbf{B} \gets \mathbf{B}_1 + \mathbf{B}_2$
    \State \Return $\mathbf{B}$
\EndFunction
\end{algorithmic}

\begin{algorithmic}[1]
\Function{NonSalientAwareQuant}{$\mathbf{W}$}
    \State $e \gets \infty$ \Comment{Searching error}
    \State $p^*_1 \gets 0$ \Comment{Optimal break-point for trisection}
    \State $p^*_2 \gets 0$ \Comment{Optimal break-point for trisection}
    \For{$i \in \text{np.linspace}(0.1, 0.9, 160)$}
        \State $p_1 \gets i \cdot \max(|\mathbf{W}|)$
        \State $p_2 \gets \sigma \times p_1$
        \If{$p_2 > 0.9 \times \max(|\mathbf{W}|)$}
            \State \textbf{continue}
        \EndIf
        \State $\mathbf{B}_1 \gets \Call{Binary}{\mathbf{W}_{|w_{i,j}| > p_2}}$
        \State $\mathbf{B}_2 \gets \Call{Binary}{\mathbf{W}_{p_1 < |w_{i,j}| \leq p_2}}$
        \State $\mathbf{B}_3 \gets \Call{Binary}{\mathbf{W}_{|w_{i,j}| \leq p_1}}$
        \If{$\|\mathbf{W} - (\mathbf{B}_1 + \mathbf{B}_2 + \mathbf{B}_3)\|^2 < e$}
            \State $e \gets \|\mathbf{W} - (\mathbf{B}_1 + \mathbf{B}_2 + \mathbf{B}_3)\|^2$
            \State $p^*_1 \gets p_1$
            \State $p^*_2 \gets p_2$
        \EndIf
    \EndFor
    \State \Return $p^*_1, p^*_2$
\EndFunction
\end{algorithmic}

\hspace{0.02cm}

\begin{algorithmic}[1]
\Function{Trisection}{$\mathbf{W}, p^*_1, p^*_2$}
    \State $\Tilde{\mathbf{B}_2} \gets \Call{Binary}{\mathbf{W}_{|w_{i,j}| > p^*_2}}$
    \State $\Tilde{\mathbf{B}_3} \gets \Call{Binary}{\mathbf{W}_{p^*_1 < |w_{i,j}| \leq p^*_2}}$
    \State $\Tilde{\mathbf{B}_4} \gets \Call{Binary}{\mathbf{W}_{|w_{i,j}| \leq p^*_1}}$
    \State \Return $\Tilde{\mathbf{B}_2}, \Tilde{\mathbf{B}_3}, \Tilde{\mathbf{B}_4}$
\EndFunction
\end{algorithmic}

\end{multicols}
\end{algorithm*}

\section{Details of Motivation Experiment}\label{appendix:motivation}

In this section, we delineate the specifics of the motivation experiments as illustrated in Figure~\ref{fig:motivation}. Initially, we elucidate the procedure for inverting the signs of elements within a matrix, as detailed in Algorithm~\ref{alg:flip}, to examine its effects on various computational tasks. This algorithm is designed to efficiently invert the signs of a specified proportion of elements in a given matrix $\mathbf{W}$. Subsequently, we employ this function on $RES\_APPROX$ and invert the signs of each binary matrix, including $B_1$ and $B_2$.

\begin{algorithm}[H]
\caption{Algorithm for Efficiently Flipping Signs of Matrix Elements}
\begin{algorithmic}[1]
\Function{FlipSignsEfficient}{$\mathbf{W}, \text{ratio}, \mathbf{C} \gets \text{None}$}
    \State $n \gets \text{numel}(\mathbf{W})$ \Comment{Total number of elements in $\mathbf{W}$}
    \State $k \gets \text{int}(n \times \text{ratio})$ \Comment{Number of elements to flip}
    
    \If{$\mathbf{C} \neq \text{None}$}
        \State \textbf{assert} $\text{shape}(\mathbf{C}) = \text{shape}(\mathbf{W})$ \Comment{Ensure $\mathbf{C}$ matches $\mathbf{W}$}
        \State $\_, \text{idx} \gets \text{sort}(\mathbf{C}.\text{view}(-1))$ \Comment{Flatten $\mathbf{C}$ and get sorted indices}
        \State $\text{idx\_to\_flip} \gets \text{idx}[:k]$ \Comment{Select least significant elements to flip}
    \Else
        \State $\text{idx\_to\_flip} \gets \text{random\_indices}(0, n, k)$ \Comment{Random select elements to flip}
    \EndIf
    
    \State $\mathbf{W_{flip}} \gets \mathbf{W}.\text{clone}()$ \Comment{Create a copy of $\mathbf{W}$}
    \State $\mathbf{W_{flip-flat}} \gets \mathbf{W_{flip}}.\text{view}(-1)$ \Comment{View the copy as a 1D tensor}
    
    \State $\mathbf{W_{flip-flat}}[\text{idx\_to\_flip}] \gets \mathbf{W_{flip-flat}}[\text{idx\_to\_flip}] \times -1$ \Comment{Flip the signs of selected elements}
    
    \State \Return $\mathbf{W_{flip}}$
\EndFunction
\end{algorithmic}
\label{alg:flip}
\end{algorithm}

\section{Details of Hardware Acceleration}\label{appendix:hardware}

\subsection{Implementation Details}

Recent advancements in low-precision computing have significantly enhanced the practical implementation of efficient neural network techniques. A prime example is the introduction of Ladder \citep{Wang2024LadderEE}, released as BitBLAS, a software library that seamlessly integrates into existing Deep Neural Network (DNN) and Large Language Model (LLM) frameworks. This integration enables highly efficient low-precision computations across various hardware platforms. The impact of these developments is evident in popular frameworks like llama.cpp \citep{ggerganov2023llama}, which now supports 1.5-bit quantization through BitNet \citep{wang2023bitnet}. This advancement has resulted in impressive performance gains, achieving 198 tokens per second on a single CPU core. Moreover, for large-scale models such as LLaMA-2-70B, the implementation of Ladder \citep{Wang2024LadderEE} to accelerate BitNet 1.58 \citep{wang2023bitnet} has yielded remarkable results, demonstrating a 4.6$\times$ speedup compared to FP16 precision.

The emergence of Sparse Tensor Cores (SPTCs) since NVIDIA's Ampere architecture has revolutionized the processing of sparse matrices, offering an efficient mechanism for handling 50\% sparsity. Theoretically, by eliminating half of the computations, SPTCs can potentially double the computational power compared to Dense Tensor Cores. There are already several research over accelerating dense tensor core, including ULPPack~\citep{Won2022ULPPACKFS}, NGEMM~\citep{Bao2019NGEMMOG} and QQQ~\citep{Zhang2024QQQQQ}. However, efficiently representing 1-bit values ($+1$, $-1$, and $0$ for sparsity) and achieving sufficient real-world acceleration pose significant challenges. To address these, we propose a novel 2-bit integer representation method, particularly useful for the General Matrix Multiply (GEMM) operation, formulated as $D = A(E) \times B + C$. Here, $A$ represents the 2:4 1-bit sparse matrix, $B$, $C$, and $D$ are dense tensors, and $E$ employs \texttt{uint16} to denote the valid indices of $A$.

Our approach introduces a 6-bit encoding scheme for each group of 2:4 sparse 1-bit values. This scheme comprises four bits for indexing and two bits for physical value representation, where $1 \rightarrow +1$, $0 \rightarrow -1$, and positions unmarked by $E$ indicate sparsity ($0$). This method significantly improves memory efficiency compared to a baseline approach using 2-bit integers to represent $-1$, $0$, and $+1$, which would require 8 bits for an equivalent group size. Consequently, our encoding method reduces memory footprint by approximately 25\%, leading to decreased global memory access requirements. In memory-bound scenarios typical of large-scale model inference, this approach theoretically offers up to a 1.333-fold increase in processing speed compared to the 2-bit variant.

\begin{figure}[t]
    \centering
    \includegraphics[width=0.75\linewidth]{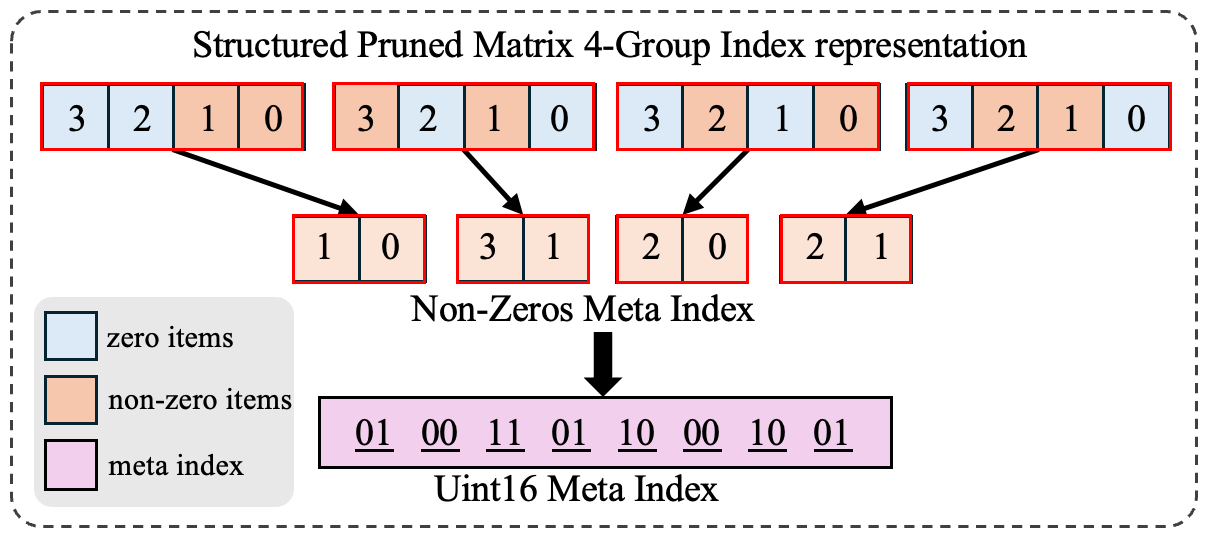}
    \caption{Structured Pruned Matrix 4-Group Index Representation for 2:4 Structured Sparsity Acceleration.}
    \label{fig:hardware-fig1}
\end{figure}

\begin{figure}[t]
    \centering
    \begin{minipage}[b]{0.57\textwidth}
        \centering
        \includegraphics[width=\linewidth]{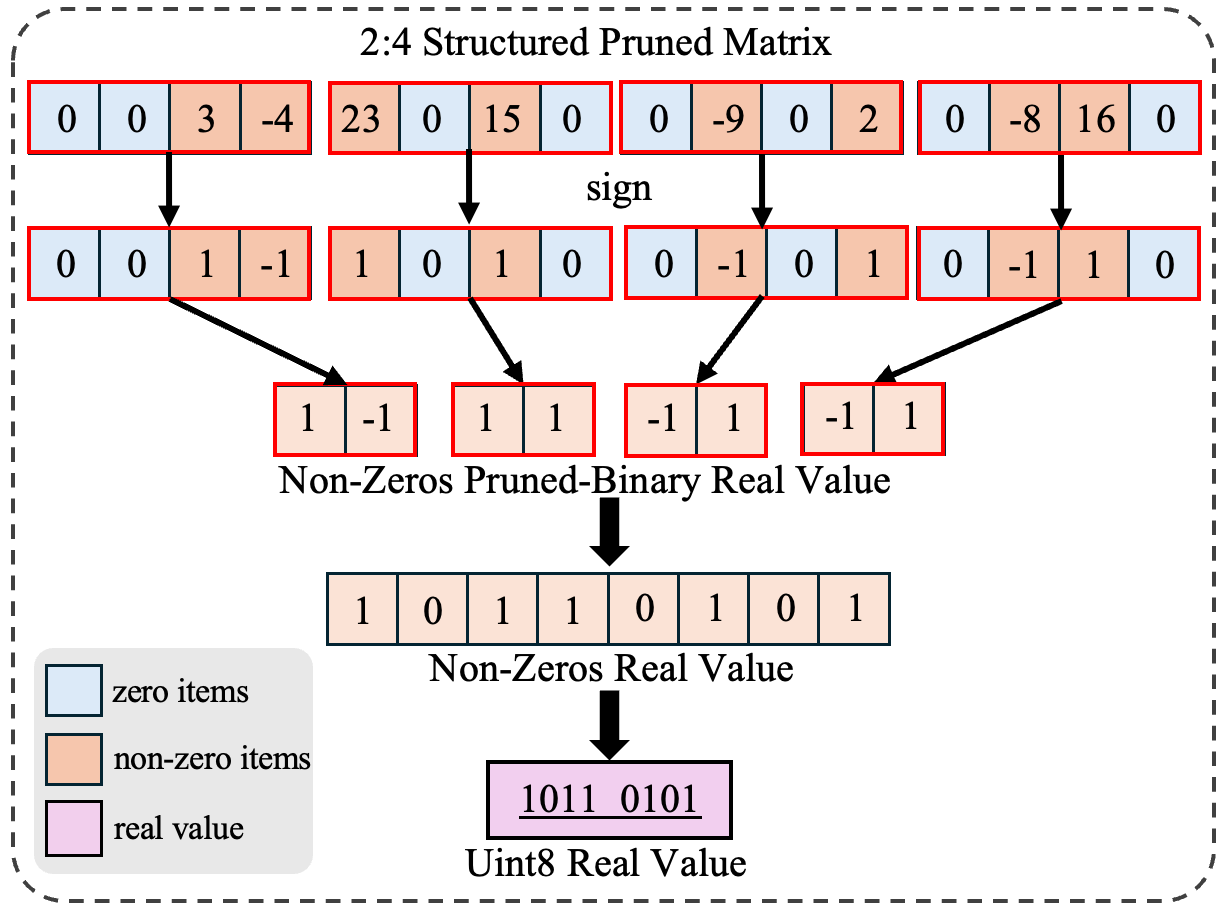}
        \caption{The overview of sparsity pattern of 1-bit kernel that convert weight matrix to structured pruned matrix. }
        \label{fig:hardware-fig2}
    \end{minipage}
    \hfill
    \begin{minipage}[b]{0.41\textwidth}
        \centering
        \includegraphics[width=\linewidth]{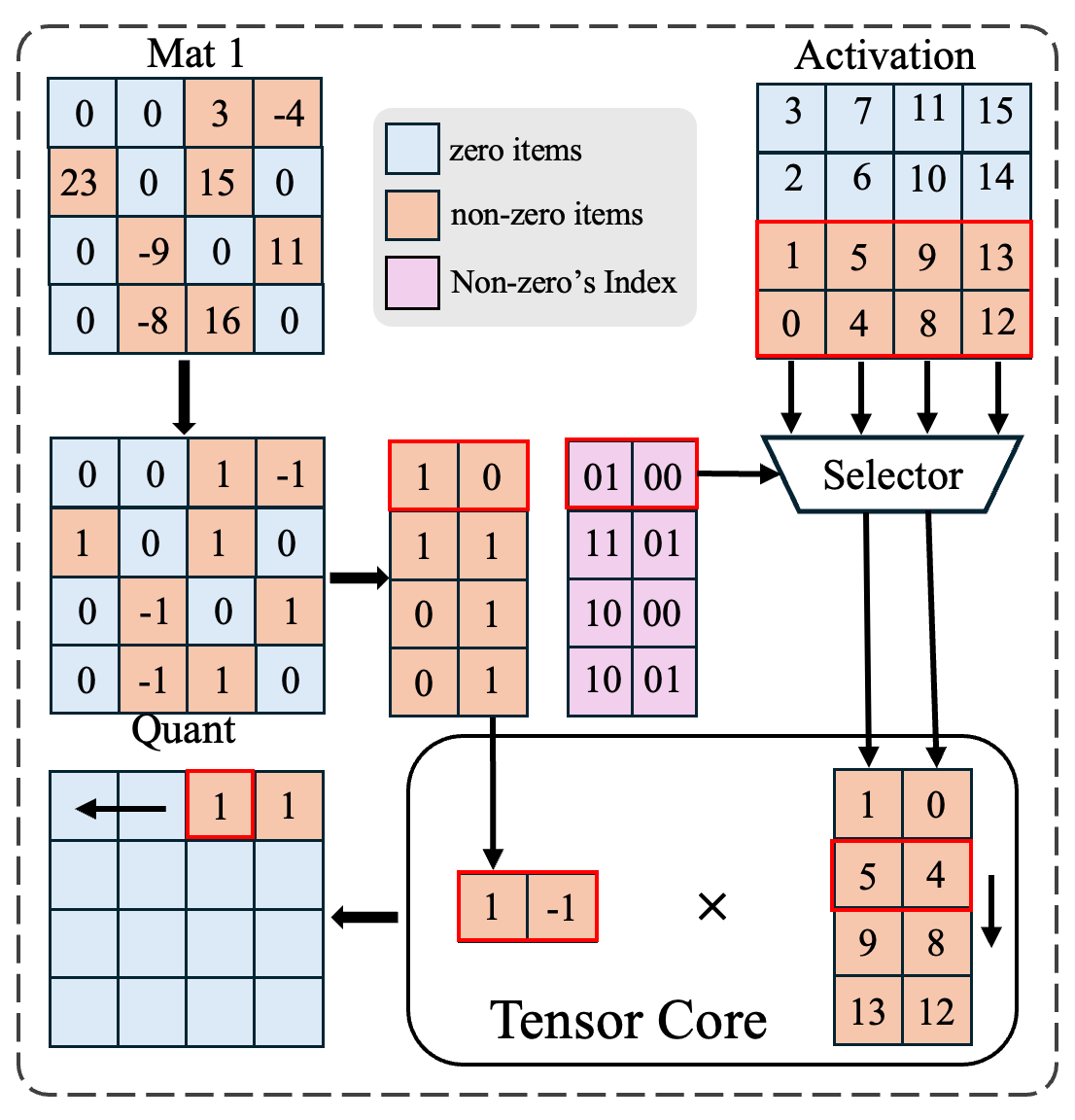}
        \caption{2:4 Structured Sparsity Matrix Multiplication Using Tensor Cores.}
        \label{fig:hardware-fig3}
    \end{minipage}
\end{figure}

To fully leverage these optimizations, we employ semi-structured pruning techniques specifically tailored for NVIDIA's GPU architecture. These techniques enable the use of Sparse Tensor Cores optimized for processing sparse matrices. By structuring the sparsity (e.g., $N$:$M$ sparsity where $N$ out of $M$ weights are non-zero), we can effectively utilize the Sparse Tensor Cores, leading to substantial improvements in processing speed and efficiency. Specifically, the process of implementing these optimizations involves several key steps in matrix compression and manipulation:

\begin{enumerate}
    \item \textbf{Matrix Compression:} The input matrix is partitioned into 4-element groups as shown in Figure~\ref{fig:hardware-fig1}. Zero elements are identified and non-zero elements are extracted. The positions of non-zero elements are recorded in a Non-Zeros Meta Index, which is then encoded into a compact \texttt{Uint16} Meta Index. This encoding facilitates efficient localization of non-zero elements during matrix operations, enhancing computational speed by enabling the omission of zero elements.

    \item \textbf{Value Compression:} Similar to matrix compression, the matrix is divided into 4-element groups as shown in Figure~\ref{fig:hardware-fig2}. The sign of each non-zero element is extracted to form the Non-Zeros Pruned-Binary Real Value. These values are then converted to a binary format, creating the Non-Zeros Real Value. Finally, the each of eight binary values are concatenated into a compact \texttt{Uint8} Real Value, optimizing storage and computation by focusing on the non-zero elements and their signs.

    \item \textbf{Matrix Multiplication with Structured Sparsity:} The input matrices undergo pruning to retain only non-zero elements and their corresponding indices as shown in Figure~\ref{fig:hardware-fig3}. The pruned matrices are then quantized, extracting non-zero values and their positions. Both processed matrices are subsequently input into the Sparse Tensor Cores, which executes efficient multiplication by focusing on the non-zero elements, resulting in a compressed and accelerated computation.
\end{enumerate}

\begin{figure}[t]
    \centering
    \includegraphics[width=0.7\linewidth]{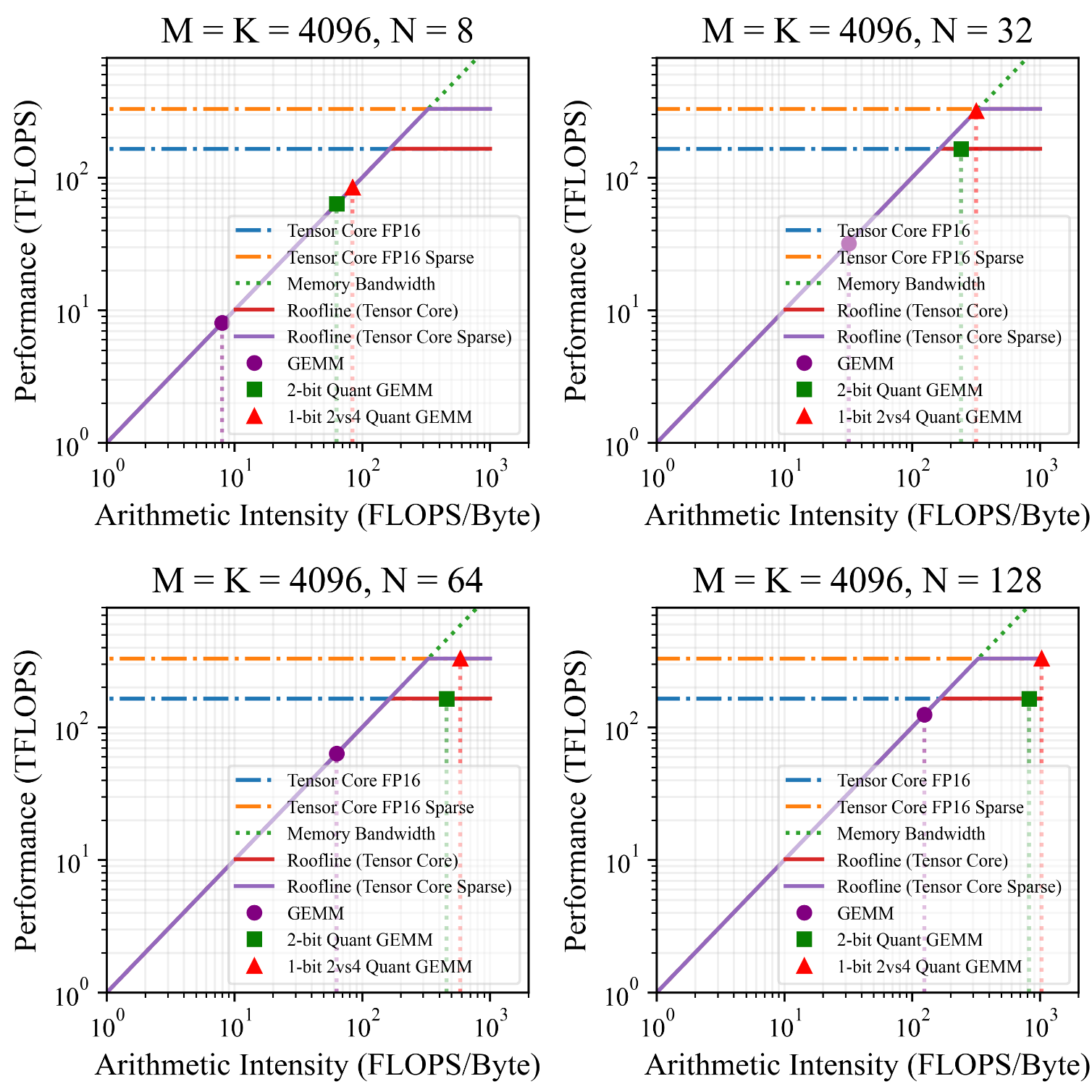}
    \caption{Roofline for Sparse GEMM Quantization.}
    \label{fig:roofline}
\end{figure}

\subsection{Theory Analysis}

To evaluate the performance of various matrix multiplication algorithms across different problem sizes, we present a comprehensive roofline model analysis in Figure~\ref{fig:roofline}. Each subplot depicts the relationship between arithmetic intensity (FLOPS/Byte) and performance (TFLOPS). 
During prefilling stage, $N$ denotes the product of sequence length and batch size. For decoding stage, $N$ denotes the batch size. $M$ and $K$ correspond to the dimension of weight matrix. 
To compare different implementations, we include standard FP16 GEMM, 2-bit quantized GEMM, and 1-bit 2vs4 quantized GEMM, alongside theoretical performance limits represented by roofline models for Tensor Core and Tensor Core Sparse operations. From our analysis, we observe that as N increases, all algorithms exhibit improved performance, with quantized versions consistently outperforming standard GEMM. We find that our 1-bit 2vs4 quantized GEMM demonstrates superior performance, particularly at larger N values, often approaching the Tensor Core Sparse roofline.

The advantages of our 1-bit 2:4 quantized GEMM kernel arise from reduced memory access overhead and the higher compute upper bound of Sparse Tensor Cores (SPTCs). When $N$ is small (particularly during the decoding phase), all GEMM kernels are memory-bound, but our 1-bit 2:4 quantized GEMM kernel achieves relatively better performance due to its higher compression rate. As  $N$ increases (especially during the prefilling stage), the quantized GEMM kernels tend to become compute-bound. In this case, our specialized GEMM kernel can theoretically achieve a 2$\times$ speedup compared to other GEMM kernels.

This extreme quantization approach significantly reduces both computational overhead and memory footprint by limiting the precision of weights to just two possible states. Such a method is particularly advantageous in resource-constrained environments, improving the deployability of large models on devices with limited hardware capabilities.

\subsection{Memory Comparison}

\begin{figure}[h]
    \centering
    \includegraphics[width=0.98\linewidth]{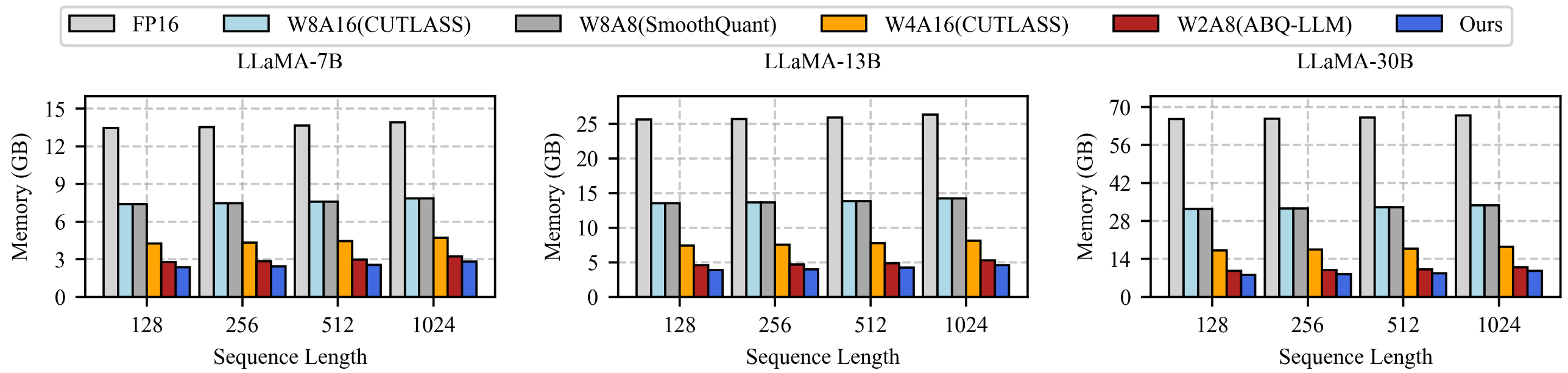}
    \caption{Memory Usage Comparison of Various Quantization Methods for LLaMA Models}
    \label{fig:memory_comparison_hardware}
\end{figure}

As illustrated in Figure~\ref{fig:memory_comparison_hardware}, we present the memory consumption of FP16, CUTLASS, ABQ-LLM, and our implementation for LLaMA-7B, 13B, and 30B models. Our proposed methodology demonstrates a substantial memory compression gain, exceeding 3.1 times that of SmoothQuant. This performance significantly surpasses current mainstream inference techniques. Furthermore, our approach achieves an approximate 15\% reduction in memory usage compared to ABQ-LLM. These notable improvements have important implications for the field of large language models (LLMs). By reducing the memory footprint, our method decreases the operational costs associated with LLM services and facilitates their practical deployment in real-world applications.

\section{Impact of Extreme Weight on the Hessian Matrix}\label{appendix:hessian}

The Hessian matrix \(H\) is defined as: $H_{ij} = \frac{\partial^2 L}{\partial w_i \partial w_j}$, where \(L\) is the loss function, and \(w_i\) and \(w_j\) are weights. If a weight \(w_k\) has extreme values, the corresponding elements in the Hessian matrix, particularly \(H_{kk}\), will be significantly larger than others.

For instance, if \(w_1\) is an extreme value, the Hessian matrix might look like:
\[
H = \begin{pmatrix}
h_{11} & h_{12} & \cdots & h_{1n} \\
h_{21} & h_{22} & \cdots & h_{2n} \\
\vdots & \vdots & \ddots & \vdots \\
h_{n1} & h_{n2} & \cdots & h_{nn}
\end{pmatrix}
\]
Here, \(h_{11}\) is much larger than other elements. This disproportionate value significantly influences the Hessian's eigenvalues, with at least one eigenvalue becoming very large.
During optimization, methods like Newton's method update weights using the inverse of the Hessian matrix:
\[
\mathbf{w}_{\text{new}} = \mathbf{w} - \eta H^{-1} \nabla L(\mathbf{w}),
\]
where \(\eta\) is the learning rate, and \(\nabla L(\mathbf{w})\) is the gradient. The presence of an extreme value in \(h_{11}\) causes the corresponding element in \(H^{-1}\) to be very small, affecting the step size in weight updates:
\[
\Delta w_1 \approx -\eta \frac{\partial L}{\partial w_1} / h_{11},
\]
\[
\Delta w_2 \approx -\eta \frac{\partial L}{\partial w_2} / h_{22}.
\]

Since \(h_{11}\) is large, \(\Delta w_1\) becomes small, indicating minimal adjustments for the extreme value weight, while \(\Delta w_2\) remains relatively larger for the normal weights.

\section{More Experimental Results}\label{appendix:ablation}

\begin{figure}[t]
    \centering
    \begin{minipage}{0.48\textwidth}
        \centering
        \includegraphics[width=\linewidth]{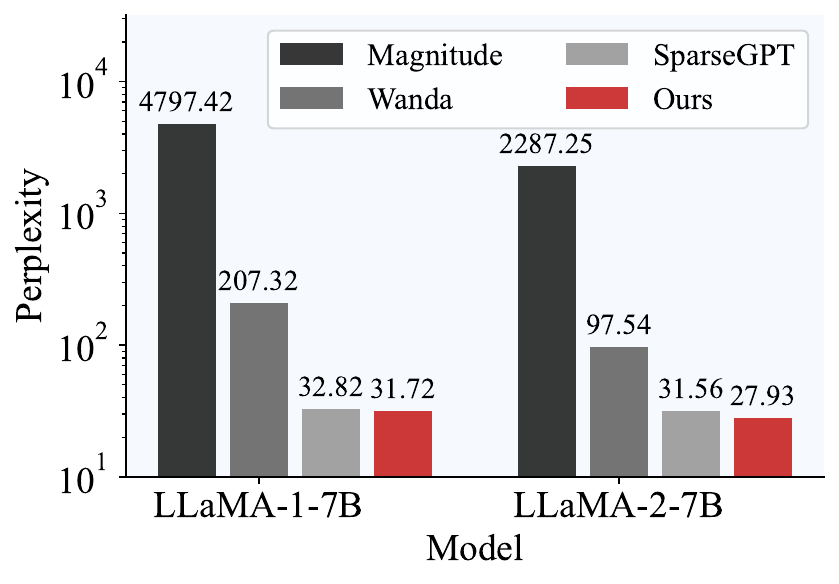}
        \caption{\textbf{Ablation study on post-training pruning metrics on STBLLM on LLaMA-1-7B and LLaMA-2-7B.} Our method achieves the best performance among these metrics. \\ \\}
        \label{fig:ablation_for_metric}
    \end{minipage}
    \hfill
    \begin{minipage}{0.48\textwidth}
        \centering
        \includegraphics[width=\linewidth]{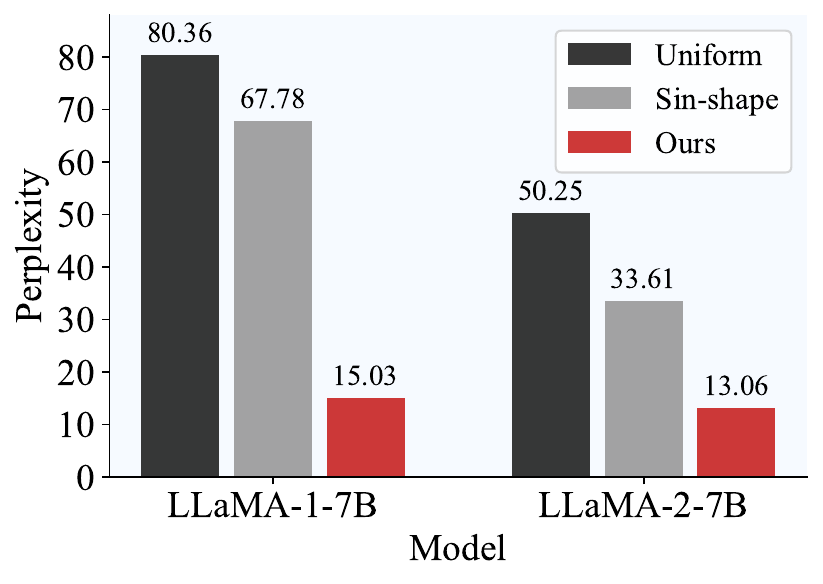}
        \caption{\textbf{Ablation study on allocation strategies on STBLLM on LLaMA-1-7B and LLaMA-2-7B.} Our strategy consistently achieves nearly identical perplexity across both models, significantly outperforming the other two allocation strategies}
        \label{fig:ablation_for_allocation_strategies}
    \end{minipage}
\end{figure}

\begin{figure}[t]
    \centering
    \begin{minipage}{1\textwidth}
        \centering
        \includegraphics[width=\linewidth]{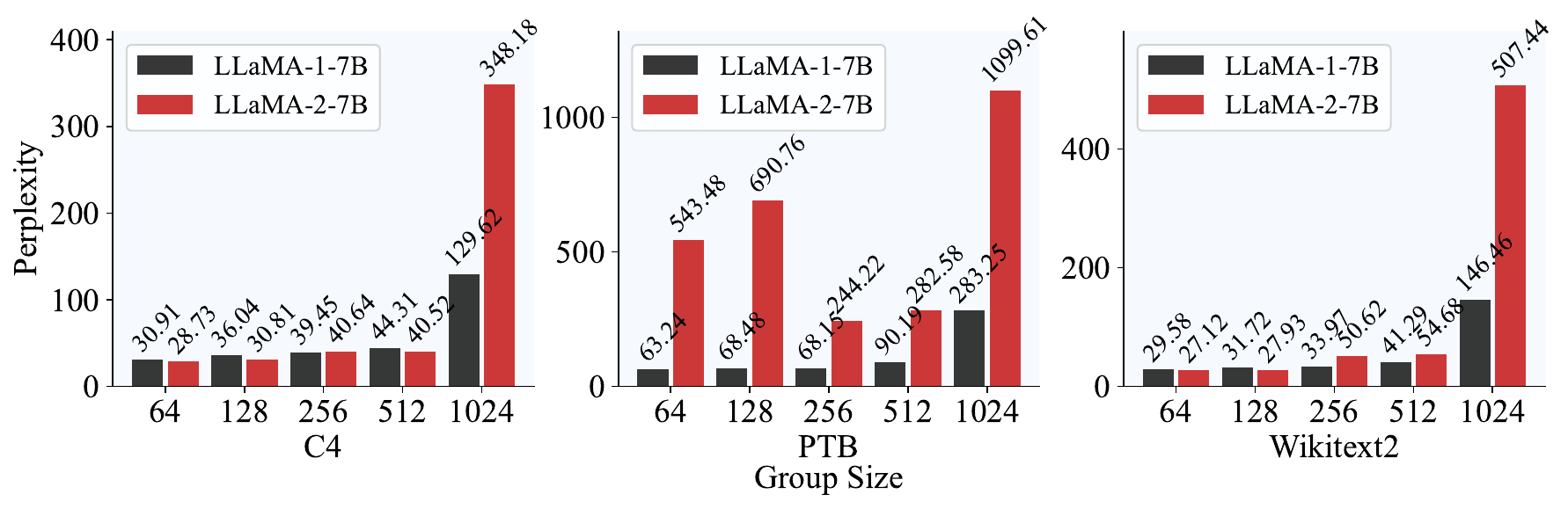}
        \caption{\textbf{Comparison across different sizes for LLaMA-1-7B and LLaMA-2-7B.}}
        \label{fig:comparison_across_different_sizes_groupsize}
    \end{minipage}
\end{figure}

\subsection{Module Ablation Study}

To evaluate the interdependent interaction between quantization and pruning within our STBLLM framework, we conduct a module ablation study. This study isolates the effects of quantization-only, pruning-only, and our combined method on the performance of the LLaMA-1-7B and LLaMA-2-7B models across the PTB, C4, and Wikitext2 datasets. The results are presented in Table \ref{tab:comparison_mo}.

\begin{table}[t]
    \centering
        \caption{Comparison of Quant-Only, Structure-Only, and ours across different datasets.}
            \resizebox{0.8\textwidth}{!}{
    \begin{tabular}{lrrrrrr}
    \toprule
    & \multicolumn{3}{c}{\textbf{LLaMA-1-7B}}  & \multicolumn{3}{c}{\textbf{LLaMA-2-7B}} \\
    \cmidrule{2-7} Dataset & Quant-Only & Structure-Only & Ours  &  Quant-Only & Structure-Only & Ours \\
    \midrule
    PTB   & 23.52 & 14.24 & 68.48   & 2071.44 & 69.25 & 690.76 \\
    C4    & 15.75 & 10.52 & 36.04   & 14.62 & 10.29 & 30.81 \\
    Wikitext2 & 12.29 & 8.13  & 31.72 & 11.17 & 7.85  & 27.93 \\
    \bottomrule
    \end{tabular}%
}
    \label{tab:comparison_mo}
\end{table}

The ablation results highlight the synergistic effect of combining quantization and pruning in our approach, significantly outperforming each method applied in isolation.

\textbf{LLaMA-1-7B Analysis}

- \textit{PTB Dataset}: Our combined method achieves a score of 68.48, markedly higher than quantization-only (23.52) and pruning-only (14.24). This demonstrates the substantial performance gains achieved by leveraging the complementary strengths of both techniques.

- \textit{C4 Dataset}: Our method scores 36.04, compared to 15.75 for quantization-only and 10.52 for pruning-only. The combined approach effectively mitigates the limitations of individual methods, resulting in superior performance.

- \textit{Wikitext2 Dataset}: The score of 31.72 for our method far exceeds the results of quantization-only (12.29) and pruning-only (8.13), underscoring the enhanced model efficiency and accuracy through our integrated approach.

\textbf{LLaMA-2-7B Analysis}

- \textit{PTB Dataset}: Although quantization-only achieves an unusually high score of 2071.44, our combined method still significantly outperforms pruning-only (690.76 vs. 69.25). This suggests that while quantization might retain certain advantageous structures, the integration with pruning leads to a more balanced and robust model.

- \textit{C4 Dataset}: The combined method's score of 30.81 surpasses quantization-only (14.62) and pruning-only (10.29), highlighting the effectiveness of our method in maintaining high performance across varying model versions.

- \textit{Wikitext2 Dataset}: Our method's score of 27.93 is higher than both quantization-only (11.17) and pruning-only (7.85), further confirming the synergistic benefits of combining these techniques.

\subsection{Ablation Study of Calibration Dataset}

Table~\ref{tab:comparison_ds} presents an ablation study comparing the performance of LLaMA-1-7B and LLaMA-2-7B models when trained on different calibration datasets: C4, PTB, and Wikitext2. The purpose of this experiment is to investigate how the choice of calibration dataset affects the models' performance on various evaluation datasets.

In this study, both LLaMA-1-7B and LLaMA-2-7B models are trained on each of the three calibration datasets separately. The trained models are then evaluated on all three datasets, resulting in a 3x3 matrix of performance scores for each model.

The performance scores in the table likely represent some evaluation metric, such as perplexity or loss, where lower values indicate better performance. The diagonal values (e.g., C4 evaluated on C4) represent in-domain performance, while off-diagonal values represent out-of-domain performance.

\begin{table}[t]
    \centering
    \caption{Comparison of C4, PTB, and Wikitext2 across LLaMA-1-7B and LLaMA-2-7B}
     \resizebox{0.66\textwidth}{!}{
    \begin{tabular}{lrrrrrr}
    \toprule
     & \multicolumn{3}{c}{\textbf{LLaMA-1-7B}} & \multicolumn{3}{c}{\textbf{LLaMA-2-7B}} \\
\cmidrule{2-7}    Dataset & C4    & PTB   & Wikitext2 & C4    & PTB   & Wikitext2 \\
    \midrule
    C4    & 36.04 & 68.48 & 31.72 & 30.81 & 690.76 & 27.93 \\
    PTB   & 54.57 & 35.13 & 49.27 & 43.04 & 4569.03 & 40.94 \\
    Wikitext2 & 40.76 & 71.81 & 20.48 & 37.01 & 1970.76 & 20.60 \\
    \bottomrule
    \end{tabular}%
    }
    \label{tab:comparison_ds}
\end{table}

\subsection{Ablation Study of Group Size}

Table~\ref{tab:size_comparison} and Figure~\ref{fig:comparison_across_different_sizes_groupsize} presents an ablation study that compares the performance of LLaMA-1-7B and LLaMA-2-7B models across different group sizes. The purpose of this experiment is to investigate how the choice of group size affects the models' performance on various evaluation datasets.

In this study, both LLaMA-1-7B and LLaMA-2-7B models are trained with different group sizes: 64, 128, 256, 512, and 1024. The trained models are then evaluated on three datasets: C4, PTB, and Wikitext2. The performance scores in the table likely represent some evaluation metric, such as perplexity or loss, where lower values indicate better performance. By comparing the performance scores across different group sizes and evaluation datasets, researchers can gain insights into the impact of group size on the models' performance and generalization capabilities.

The results show that the performance of both models varies with the choice of group size. For LLaMA-1-7B, the best performance on C4 and Wikitext2 is achieved with a group size of 64, while for PTB, the best performance is obtained with a group size of 128. For LLaMA-2-7B, the best performance on C4 and Wikitext2 is also achieved with a group size of 64, while for PTB, the best performance is obtained with a group size of 256. Interestingly, the performance of both models deteriorates significantly when the group size is increased to 1024, suggesting that excessively large group sizes may lead to overfitting or other training issues.

\begin{table}[t]
    \centering
    \caption{Comparison across different sizes for LLaMA-1-7B and LLaMA-2-7B}
    \begin{adjustbox}{max width=\textwidth}
    \begin{tabular}{ccccccc}
        \toprule
        \multicolumn{1}{c}{ } & \multicolumn{3}{c}{\textbf{LLaMA-1-7B}} & \multicolumn{3}{c}{\textbf{LLaMA-2-7B}} \\
        \cmidrule{2-7} Group Size
        & C4 & PTB & Wikitext2 & C4 & PTB & Wikitext2 \\
        \midrule
        64 & 30.91 & 63.24 & 29.58 & 28.73 & 543.48 & 27.12 \\
        128 & 36.04 & 68.48 & 31.72 & 30.81 & 690.76 & 27.93 \\
        256 & 39.45 & 68.15 & 33.97 & 40.64 & 244.22 & 50.62 \\
        512 & 44.31 & 90.19 & 41.29 & 40.52 & 282.58 & 54.68 \\
        1024 & 129.62 & 283.25 & 146.46 & 348.18 & 1099.61 & 507.44 \\
        \bottomrule
    \end{tabular}
    \end{adjustbox}
    \label{tab:size_comparison}
\end{table}

\begin{table}[t]
\centering
\caption{Motivation vs Top Percentage}
\begin{tabular}{cc|cc|cc}
\toprule
\textbf{Top Percentage} & \textbf{Perplexity} & \textbf{Top Percentage} & \textbf{Perplexity} & \textbf{Top Percentage} & \textbf{Motivation} \\
\midrule
0.01 & 27.770422 & 0.02 & 30.168285 & 0.03 & 34.049734 \\
0.04 & 36.191769 & 0.05 & 33.821476 & 0.06 & 36.452296 \\
0.07 & 38.702617 & 0.08 & 39.169894 & 0.09 & 44.818825 \\
0.10 & 54.451229 & 0.11 & 49.835159 & 0.12 & 71.762848 \\
0.13 & 52.129317 & 0.14 & 52.568348 & 0.15 & 65.945448 \\
0.16 & 62.712751 & 0.17 & 117.990227 & 0.18 & 138.912356 \\
\bottomrule
\end{tabular}

\label{table:motivation_top_Percentage}
\end{table}

The provided Figure~\ref{fig:motivation} and  Table~\ref{table:motivation_top_Percentage} present an experiment that investigates the relationship between the top Percentage of data and the corresponding perplexity scores in a LM. The purpose of this experiment is to understand how the choice of top Percentage affects the model's performance and to determine an optimal threshold for data selection.
In this experiment, we randomly flip 1\%-16\% weights from binarized LM and evaluate their downstream tasks' performance including ARC~\citep{clark2018think_arc}, BoolQ~\citep{OpenBookQA2018}, Hellaswag~\citep{zellers2019hellaswag} and RTE~\citep{Chakrabarty2021FigurativeLI_RTE}. Table~\ref{table:motivation_top_Percentage} shows the perplexity scores for each top Percentage. Lower perplexity scores indicate better language model performance, as the model is better able to predict the next word in a sequence. 

Figure~\ref{fig:motivation} provides a visual representation of the relationship between the top Percentage and perplexity scores. It shows that the perplexity scores initially improve as the top Percentage increases, indicating that including more high-quality data points benefits the model's performance. However, beyond a certain threshold (around 0.05 to 0.10), the perplexity scores start to deteriorate, suggesting that including lower-quality data points negatively impacts the model's performance.

\end{document}